\documentclass[11pt,a4paper]{article}
\usepackage[acceptedWithA]{tacl2018v2}
\usepackage{times,latexsym}
\usepackage{url}
\usepackage[T1]{fontenc}
\usepackage{bbm}
\usepackage{appendix}

\usepackage{xspace,mfirstuc,tabulary}

\newif\iftaclinstructions
\taclinstructionsfalse
\iftaclinstructions
\renewcommand{\confidential}{}
\renewcommand{\anonsubtext}{(No author info supplied here, for consistency with
TACL-submission anonymization requirements)}
\newcommand{\instr}
\fi

\iftaclpubformat

\else

\fi

\makeatletter
\newcommand\footnoteref[1]{\protected@xdef\@thefnmark{\ref{#1}}\@footnotemark}
\makeatother
\renewcommand{\cite}[2][]{\citep[#1]{#2}}
\renewcommand{\newcite}[2][]{\citet[#1]{#2}}
\newcommand{\citeposs}[1]{\citeauthor{#1}'s \citeyearpar{#1}}

\usepackage[font=small,labelfont=bf]{caption}
\usepackage{amsmath,mathrsfs,harpoon}
\usepackage{bbding}
\usepackage{wasysym}
\usepackage{multirow}
\usepackage{url}
\usepackage{graphicx}
\usepackage{amsfonts}
\usepackage{amssymb}
\usepackage{fixltx2e}
\usepackage{enumitem}
\usepackage{color}
\usepackage{mathtools}
\usepackage{tikz}
\usepackage{tikz-dependency}
\pgfdeclarelayer{bg}
\pgfsetlayers{bg,depgroups,main}
\usetikzlibrary{backgrounds}
\usetikzlibrary{automata,positioning,arrows.meta,decorations.pathmorphing,decorations.pathreplacing,calc}
\usepackage{amsopn}
\usepackage{soul}
\usepackage{algorithmicx}
\usepackage{algorithm}
\usepackage{accents}
\usepackage[noend]{algpseudocode}
\usepackage{util}
\usepackage{CJKutf8}
\usepackage{CJK}

\newcommand{\cutforspace}[1]{}
\usepackage[disable]{todonotes}

\newlength{\extramargin}
\makeatletter\if@todonotes@disabled\else
\usepackage{calc}
\usepackage[paperwidth=\paperwidth+\extramargin*2,marginparwidth=\marginparwidth+\extramargin,width=\textwidth,height=\textheight]{geometry}
\fi

\usepackage[linguistics]{forest}
\forestset{pad/.style={minimum width=5em},
    empty nodes/.style={
            delay={where content={}{shape=coordinate,for parent={
                  for children={anchor=north}}}{}}
}}

\DeclareMathOperator*{\argmin}{argmin}
\definecolor{undp}{RGB}{128,255,128}
\definecolor{undr}{RGB}{0,100,0}
\definecolor{undb}{RGB}{0,0,255}
\newcommand{\litf}[1]{\mbox{\textnormal{\texttt{#1}}}}

\newcommand{\undr}[1]{{\litf{\color{undr}#1}}}
\newcommand{\undb}[1]{{\litf{\color{undb}#1}}}

\newcommand{\posf}[1]{\texttt{#1}}
\newcommand{\depf}[1]{\texttt{#1}}
\setlist[enumerate]{itemsep=0mm}
\newlength\min@xx

\makeatother

\usepackage{xspace}
\newcommand{\bos}{\texttt{BOS}\xspace}
\newcommand{\eos}{\texttt{EOS}\xspace}

\newcommand{\ra}[1]{\hspace{-3pt}\mathrel{\raisebox{-3pt}{$\xrightarrow{#1}$}}\hspace{-3pt}}
\newcommand{\myrarw}{\hspace{-2pt}\mapsto\hspace{-2pt}}

\newcommand{\appref}[1]{\cref{#1}}

\makeatletter
\newcommand{\printfnsymbol}[1]{%
  \textsuperscript{\@fnsymbol{#1}}%
}
\makeatother

\title{A Generative Model for Punctuation in Dependency Trees}

\author{Xiang Lisa Li\thanks{~~Equal contribution.}\and Dingquan Wang\printfnsymbol{1}\and Jason Eisner  \\
        Department of Computer Science, Johns Hopkins University \\
        \texttt{xli150@jhu.edu, \{wdd,jason\}@cs.jhu.edu}}

\date{}

\begin{document}
\maketitle
\pagestyle{plain}
\thispagestyle{plain}
\setcounter{page}{357}
\begin{abstract}
    Treebanks traditionally treat punctuation marks as ordinary words, but linguists have suggested that a tree's ``true'' punctuation marks are not observed \cite{nunberg1990linguistics}.  These latent ``underlying'' marks serve to delimit or separate constituents in the syntax tree.  When the tree's yield is rendered as a written sentence, a string rewriting mechanism transduces the underlying marks into ``surface'' marks, which are part of the observed (surface) string but should not be regarded as part of the tree.  We formalize this idea in a generative model of punctuation that admits efficient dynamic programming. We train it without observing the underlying marks, by locally maximizing the incomplete data likelihood (similarly to the EM algorithm).  When we use the trained model to reconstruct the tree's underlying punctuation, the results appear plausible across 5 languages, and in particular are consistent with Nunberg's analysis of English.  We show that our generative model can be used to beat baselines on punctuation restoration. Also, our reconstruction of a sentence's underlying punctuation lets us appropriately render the surface punctuation (via our trained underlying-to-surface mechanism) when we syntactically transform the sentence.
\end{abstract}

\vspace{-12pt}
\section{Introduction}
\label{sec:intro}

Punctuation enriches the expressiveness of written language.
When converting from spoken to written language, punctuation indicates pauses or pitches; expresses propositional attitude; and is conventionally associated with certain syntactic constructions such as apposition, parenthesis, quotation, and conjunction.

In this paper, we present a latent-variable model of punctuation usage, inspired by the rule-based approach to English punctuation of \citet{nunberg1990linguistics}.  
Training our model on English data learns rules that are consistent with Nunberg’s hand-crafted rules.  
Our system is automatic, so we use it to obtain rules for Arabic, Chinese, Spanish, and Hindi as well. 

Moreover, our rules are stochastic, which allows us to reason probabilistically about ambiguous or missing punctuation.
 Across the 5 languages, our model predicts surface punctuation better than baselines, as measured both by perplexity (\cref{sec:perplexity}) and by accuracy on a punctuation restoration task (\cref{sec:restore}).  We also use our model to correct the punctuation of non-native writers of English (\cref{sec:corr}), and to maintain natural punctuation style when syntactically transforming English sentences (\cref{sec:gd}).  In principle, our model could also be used within a generative parser, allowing the parser to evaluate whether a candidate tree truly explains the punctuation observed in the input sentence (\cref{sec:future}).

\paragraph{Punctuation is interesting} 
In {\it The Linguistics of Punctuation}, \citet{nunberg1990linguistics} argues that punctuation (in English) is more than a visual counterpart of spoken-language prosody, but forms a linguistic system that involves ``interactions of point indicators (i.e. commas, semicolons, colons, periods and dashes).'' He proposes that much as in phonology \cite{ChomskyHalle68},
a grammar generates \defn{underlying} punctuation which then transforms into the observed \defn{surface} punctuation.

Consider generating a sentence from a syntactic grammar as follows:
\begin{quote}
  \scriptsize
  \litf{Hail the king [\undb{,} Arthur Pendragon \undb{,}]}
  \litf{\ \ \ \ [\undb{,} who wields [ \undb{``} Excalibur \undb{''} ] \undb{,}] \undb{.}}
\end{quote}
Although the full tree is not depicted here, some of the constituents are indicated with brackets.  In this underlying generated tree, each appositive NP is surrounded by commas.  On the surface, however, the two adjacent commas after \litf{Pendragon} will now be collapsed into one, and the final comma will be absorbed into the adjacent period.  Furthermore, in American English, the typographic convention is to move the final punctuation inside the quotation marks.  Thus a reader sees only this modified surface form of the sentence: 
\begin{quote}
  \scriptsize
  \litf{Hail the king\undr{,} Arthur Pendragon\undr{,}}
  \litf{\ \ \ \ who wields \undr{``}Excalibur\undr{.''}}
\end{quote}
Note that these modifications are \emph{string} transformations that do not see or change the tree.
The resulting surface punctuation marks may be clues to the parse tree, but (contrary to NLP convention) they should not be included as nodes in the parse tree.  Only the underlying marks play that role.

\paragraph{Punctuation is meaningful} 
\citet{W02-1011} use question and exclamation marks as clues to sentiment.  Similarly, quotation marks may be used to mark titles, quotations, reported speech, or dubious terminology
\cite{university2010chicago}.  Because of examples like this, methods for determining the similarity or meaning of syntax trees, such as a tree kernel \cite{W11-0705} or a recursive neural network \cite{P15-1150}, should ideally be able to consider where the underlying punctuation marks attach.

\paragraph{Punctuation is helpful} 
Surface punctuation remains correlated with syntactic phrase structure.  NLP systems for generating or editing text must be able to deploy surface punctuation as human writers do.  Parsers and grammar induction systems benefit from the presence of surface punctuation marks \cite{C94-1069,Spitkovsky:2011:PMP:2018936.2018939}. It is plausible that they could do better with a linguistically informed model that explains exactly \emph{why} the surface punctuation appears where it does.  Patterns of punctuation usage can also help identify the writer's native language \citep{markov-nastase-strapparava:2018:C18-1}.

\paragraph{Punctuation is neglected}
Work on syntax and parsing tends to treat punctuation as an afterthought rather than a phenomenon governed by its own linguistic principles.
Treebank annotation guidelines for punctuation tend to adopt simple heuristics like ``attach to the highest possible node that preserves projectivity'' \cite{bies1995bracketing,nivre-ud-guide}.\footnote{\url{http://universaldependencies.org/u/dep/punct.html}}
Many dependency parsing works exclude punctuation from evaluation \cite{nivre2007maltparser,P10-1001,D14-1082,lei-EtAl:2014:P14-1,Q16-1023}, although some others retain punctuation \cite{nivre-EtAl:2007:EMNLP-CoNLL2007,goldberg-elhadad:2010:NAACLHLT,dozat2017}.

In tasks such as word embedding induction \cite{mikolov2013,D14-1162} and machine translation \cite{zens2002phrase}, punctuation marks are usually either removed or treated as ordinary words \cite{rehurek_lrec}.%
\cutforspace{However, punctuation marks are special: they often come in matched pairs, and they are subject to orthographic rewrite rules as discussed above.}%

Yet to us, building a parse tree on a \emph{surface} sentence seems as inappropriate as morphologically segmenting a \emph{surface} word.  In both cases, one should instead analyze the latent \emph{underlying} form, jointly with recovering that form.  For example, the proper segmentation of English \litf{hoping} is not \litf{hop-ing} but \litf{hope-ing} (with underlying \litf{e}), and the proper segmentation of \litf{stopping} is neither \litf{stopp-ing} nor \litf{stop-ping} but \litf{stop-ing} (with only one underlying \litf{p}). \newcite{cotterell-peng-eisner-2015,cotterell-vieira-schuetze-2016} get this right for morphology.  We attempt to do the same for punctuation.

\section{Formal Model}
\label{sec:formal}
\newcommand{\bossym}{\raisebox{-0pt}{\^{}}}
\newcommand{\p}{p}
\newcommand{\dtree}{T}
\newcommand{\utree}{\dtree'}
\newcommand{\struct}{D}
\newcommand{\utok}{\upsilon}
\newcommand{\uslot}{u}
\newcommand{\sslot}{x}
\newcommand{\upunc}{\vec{\uslot}}
\newcommand{\spunc}{\vec{\sslot}}
\newcommand{\dsent}{\bar{\vec{d}}}
\newcommand{\usent}{\bar{\upunc}}
\newcommand{\ssent}{\bar{\spunc}}
\newcommand{\node}{w}
\newcommand{\lpuncteme}{l}
\newcommand{\rpuncteme}{r}
\newcommand{\ATT}{\textsc{Attach}}
\newcommand{\NC}{\textsc{NoisyChannel}}
\newcommand{\psyn}{p_{\mathrm{syn}}}
\newcommand{\attp}{\vec{\theta}}
\newcommand{\ncp}{\vec{\phi}}
\setlength{\fboxsep}{.8pt}
\begin{figure}
\scalebox{.93}{
\hspace{-.5em}\begin{dependency}[label style={inner sep=0ex,outer sep=0ex,fill=yellow}, edge below, edge slant=5pt, edge vertical padding=-1pt]
\tikzstyle{every node}=[font=\scriptsize]
\begin{deptext}[column sep=.05em]
\&[-2pt]\litf{Dale}\&\&[7pt]\litf{means}\&\&[-2pt]\litf{river}\&\litf{valley}\&[7pt]\&\\
\end{deptext}
\deproot[edge unit distance=.775ex]{4}{\raisebox{1pt}{\depf{root}}}
\depedge[edge unit distance=.95ex]{4}{2}{\raisebox{1pt}{\depf{nsubj}}}
\depedge[edge unit distance=.8ex,label style={above,draw=none,fill=none}]{7}{6}{}
\depedge[edge unit distance=.625ex]{4}{7}{\raisebox{1pt}{\depf{dobj}}}
\begin{pgfonlayer}{bg}
    \node[draw=none,text width=4cm] at ($(\wordref{1}{1})+(-1.1, -.3)$) {\scriptsize Unpunctuated Tree:\hspace{6pt}$\dtree$};
\draw[dashed,rounded corners=2pt] ($(\wordref{1}{1})+(-3.3, .15)$) rectangle ($(\wordref{1}{9})+(-.15, -.75)$);
\end{pgfonlayer}
\end{dependency}
} \vspace{-18pt}

\scalebox{.93}{
\hspace{-.5em}\begin{dependency}[label style={inner sep=0ex,outer sep=0ex,fill=yellow}, edge slant=5pt, edge vertical padding=-2pt]
\tikzstyle{every node}=[font=\scriptsize]
\begin{deptext}[column sep=.05em, row sep=1em]
    \litf{\undb{``}}\&[-4pt]\litf{Dale}\&[-4pt]\litf{\undb{''}}\&[6pt]\litf{means}\&[-2.5pt]\litf{\undb{``}}\&[-4pt]\litf{river}\&[0pt]\litf{valley}\&[-4pt]\litf{\undb{''}}\&[-4pt]\litf{\undb{.}}\\
    \litf{\undr{``}}\&[-4pt]\litf{Dale}\&[-4pt]\litf{\undr{''}}\&[6pt]\litf{means}\&[-2.5pt]\litf{\undr{``}}\&[-4pt]\litf{river}\&[0pt]\litf{valley}\&[-4pt]\litf{\undr{.}}\&[-4pt]\litf{\undr{''}}\\
\end{deptext}
\deproot[edge unit distance=1.1ex]{4}{\raisebox{1pt}{\depf{root}}}
\node[draw=none,text width=4cm] at ($(\rootref)+(2.2, -.05)$) {\undb{.}};
\draw [->,decorate,decoration={snake,pre length=1.5pt,post length=1.5pt,amplitude=1pt,segment length=4pt}] ($(\wordref{1}{4})+(0, 1.45)$) -- ($(\wordref{1}{4})+(0, .9)$);
\depedge[edge unit distance=.95ex]{4}{2}{\raisebox{1pt}{\depf{nsubj}}}
\storelabelnode\nsubjlab
\node[draw=none,text width=4cm] at ($(\nsubjlab)+(1.545, 0)$) {\undb{``}\hspace{16.5pt}\undb{''}};
\draw [->,decorate,decoration={snake,pre length=1.5pt,post length=1.5pt,amplitude=1pt,segment length=4pt}] ($(\wordref{1}{3})+(.12, 1.45)$) -- ($(\wordref{1}{3})+(.12, .65)$);
\depedge[edge unit distance=1ex,label style={above,draw=none,fill=none}]{7}{6}{}
\depedge[edge unit distance=.625ex]{4}{7}{\raisebox{1pt}{\depf{dobj}}}
\storelabelnode\dobjlab
\node[draw=none,text width=4cm] at ($(\dobjlab)+(1.605, 0)$) {\undb{``}\hspace{13.2pt}\undb{''}};
\draw [->,decorate,decoration={snake,pre length=1.5pt,post length=1.5pt,amplitude=1pt,segment length=4pt}] ($(\wordref{1}{5})+(.575, 1.45)$) -- ($(\wordref{1}{5})+(.575, .65)$);
\begin{pgfonlayer}{bg}
\node[draw=none,text width=4cm] at ($(\wordref{1}{1})+(-1.1, 1.15)$) {\ATT};
\draw[fill=blue!20,draw=none,rounded corners=2pt] ($(\wordref{1}{1})+(-3.2, -.2)$) rectangle ($(\wordref{1}{9})+(.15, .95)$);
\node[draw=none,text width=4cm] at ($(\wordref{1}{1})+(.3, .6)$) {tree:\hspace{18pt}$\utree$};
\node[draw=none,text width=4cm] at ($(\wordref{1}{1})+(-1.1, .25)$) {Underlying};
\node[draw=none,text width=4cm] at ($(\wordref{1}{1})+(.3, .25)$) {sequence:{\color{undb}\hspace{1pt}$\upunc$}};
\node[draw=none,text width=4cm] at ($(\wordref{1}{1})+(.3, 0)$) {sentence:\hspace{2pt}\hspace{1pt}$\usent$};
\draw[fill=green!20,draw=none,rounded corners=2pt] ($(\wordref{2}{1})+(-3.2, -.4)$) rectangle ($(\wordref{2}{9})+(.15, .2)$);
\node[draw=none,text width=4cm] at ($(\wordref{2}{1})+(-1.1, -.1375)$) {Surface};
\node[draw=none,text width=4cm] at ($(\wordref{2}{1})+(.3, 0)$) {sentence:\hspace{3.5pt}$\ssent$};
\node[draw=none,text width=4cm] at ($(\wordref{2}{1})+(.3, -.275)$) {sequence:{\color{undr}\hspace{2pt}$\spunc$}};
\node[draw=none,text width=4cm] at ($(\wordref{2}{1})+(-1.1, .45)$) {\NC};
\node[draw=none] at ($(\wordref{1}{1})+(.02, .25)$) {\color{undb}$\uslot_0$};
\node[draw=none] at ($(\wordref{1}{3})+(.02, .25)$) {\color{undb}$\uslot_1$};
\node[draw=none] at ($(\wordref{1}{5})+(.02, .25)$) {\color{undb}$\uslot_2$};
\node[draw=none] at ($(\wordref{1}{6})+(.52, .25)$) {\color{undb}$\uslot_3$};
\node[draw=none] at ($(\wordref{1}{8})+(.12, .25)$) {\color{undb}$\uslot_4$};
\node[draw=none] at ($(\wordref{2}{1})+(.02, -.275)$) {\color{undr}$\sslot_0$};
\node[draw=none] at ($(\wordref{2}{3})+(.02, -.275)$) {\color{undr}$\sslot_1$};
\node[draw=none] at ($(\wordref{2}{5})+(.02, -.275)$) {\color{undr}$\sslot_2$};
\node[draw=none] at ($(\wordref{2}{6})+(.52, -.275)$) {\color{undr}$\sslot_3$};
\node[draw=none] at ($(\wordref{2}{8})+(.12, -.275)$) {\color{undr}$\sslot_4$};
\draw[rounded corners=2pt] ($(\wordref{1}{1})+(-.1, -.2)$) rectangle ($(\wordref{1}{1})+(.1,.15)$);
\draw[rounded corners=2pt] ($(\wordref{1}{3})+(-.1, -.2)$) rectangle ($(\wordref{1}{3})+(.1,.15)$);
\draw[rounded corners=2pt] ($(\wordref{1}{5})+(-.1, -.2)$) rectangle ($(\wordref{1}{5})+(.1,.15)$);
\draw[rounded corners=2pt] ($(\wordref{1}{8})+(-.1, -.2)$) rectangle ($(\wordref{1}{8})+(.3,.15)$);
\draw[rounded corners=2pt] ($(\wordref{2}{1})+(-.1, -.15)$) rectangle ($(\wordref{2}{1})+(.1,.2)$);
\draw[rounded corners=2pt] ($(\wordref{2}{3})+(-.1, -.15)$) rectangle ($(\wordref{2}{3})+(.1,.2)$);
\draw[rounded corners=2pt] ($(\wordref{2}{5})+(-.1, -.15)$) rectangle ($(\wordref{2}{5})+(.1,.2)$);
\draw[rounded corners=2pt] ($(\wordref{2}{8})+(-.1, -.15)$) rectangle ($(\wordref{2}{8})+(.3,.2)$);
\draw [->,decorate,decoration={snake,pre length=1.5pt,post length=1.5pt,amplitude=1pt,segment length=4pt}] ($(\wordref{1}{1})+(0, -.2)$) -- ($(\wordref{2}{1})+(0, .2)$);
\draw [->,decorate,decoration={snake,pre length=1.5pt,post length=1.5pt,amplitude=1pt,segment length=4pt}] ($(\wordref{1}{3})+(0, -.2)$) -- ($(\wordref{2}{3})+(0, .2)$);
\draw [->,decorate,decoration={snake,pre length=1.5pt,post length=1.5pt,amplitude=1pt,segment length=4pt}] ($(\wordref{1}{5})+(0, -.2)$) -- ($(\wordref{2}{5})+(0, .2)$);
\draw [->,decorate,decoration={snake,pre length=1.5pt,post length=1.5pt,amplitude=1pt,segment length=4pt}] ($(\wordref{1}{8})+(0.115, -.2)$) -- ($(\wordref{2}{8})+(0.115, .2)$);
\end{pgfonlayer}
\end{dependency}
}
\caption{\label{fig:gen_story}The generative story of a sentence.
\cutforspace{(the node labelled by \depf{compound} from \litf{valley} to \litf{river} is suppressed for readability)}%
Given an unpunctuated tree $\dtree$ at top, at each node $\node \in \dtree$, the \ATT{} process stochastically attaches a {\color{undb}left puncteme $\lpuncteme$} and a {\color{undb}right puncteme $\rpuncteme$}, which may be empty. 
The resulting tree $\utree$ has underlying punctuation $\upunc$. Each slot's punctuation {\color{undb}$\uslot_i\in\upunc$} is rewritten to {\color{undr}$\sslot_i\in\spunc$} by \NC{}.}
\end{figure}

We propose a probabilistic generative model of sentences (\cref{fig:gen_story}):
\begin{align}\label{eqn:model}
\p(\ssent) = \textstyle{\sum_{\dtree,\utree}} \psyn(\dtree) \cdot p_\attp(\utree \!\mid\! \dtree) \cdot p_\ncp(\ssent \!\mid\! \usent(\utree)) \raisetag{30pt}
\end{align}
First, an \emph{unpunctuated} dependency tree $\dtree$ is stochastically generated by some recursive process $\psyn$ \citep[e.g.,][Model C]{eisner-1996-coling}.\footnote{Our model could be easily adapted to work on constituency trees instead.}  
Second, each constituent (i.e., dependency subtree) sprouts optional underlying punctuation at its left and right edges, according to a probability distribution $p_\attp$ that depends
on the constituent's syntactic role (e.g., \depf{dobj} for ``direct object'').
This \emph{punctuated} tree $\utree$ yields the underlying string $\usent = \usent(\utree)$, which is edited by a finite-state noisy channel $p_\ncp$ to arrive at the surface sentence $\ssent$.  

This third step may alter the sequence of punctuation tokens at each \defn{slot} between words---for example, in \cref{sec:intro}, collapsing the double comma \undb{, ,} between \litf{Pendragon} and \litf{who}.
$\upunc$ and $\spunc$ denote just the punctuation at the slots of $\usent$ and $\ssent$ respectively, with $\uslot_i$ and $\sslot_i$ denoting the punctuation token sequences at the $i$\textsuperscript{th} slot.  Thus, the transformation at the $i$\textsuperscript{th} slot is $\uslot_i \mapsto \sslot_i$.

Since this model is generative, we could train it without any supervision to explain the observed surface string $\ssent$: maximize the likelihood $p(\ssent)$ in \eqref{eqn:model}, marginalizing out the possible $\dtree,\utree$ values.

In the present paper, however, we exploit known $\dtree$ values (as observed in the ``depunctuated'' version of a treebank).  Because $\dtree$ is observed, we can jointly train $\attp,\ncp$ to maximize just
\begin{equation}\label{eqn:conditmodel}
p(\spunc \mid \dtree) = \sum_{\utree} p_\attp(\utree \mid \dtree) \cdot p_\ncp(\spunc \mid \upunc(\utree))
\end{equation}
That is, the $\psyn$ model that generated $\dtree$ becomes irrelevant, but we still try to predict what surface
punctuation will be added to $\dtree$.
We still marginalize over the underlying punctuation marks $\upunc$.  These are \emph{never observed}, but they must explain the surface punctuation marks $\spunc$ (\cref{sec:nc}), and they must be explained in turn by the syntax tree $\dtree$ (\cref{sec:gen}).  The trained generative model then lets us restore or correct punctuation in new trees $\dtree$ (\cref{sec:app}).

\subsection{Generating Underlying Punctuation}\label{sec:gen}
\newcommand{\wlw}{w}
\newcommand{\wl}{\vec{\wlw}}
\newcommand{\ltr}{L}
\newcommand{\vf}{\vec{f}}
\newcommand{\pos}{g}
\newcommand{\dep}{d}
\newcommand{\punctemes}{\mathcal{V}}   
\newcommand{\punctemepairs}{\mathcal{W}}
\newcommand{\tags}{\punctemes^*}
\newcommand{\DIST}{\operatorname{D}}
\newcommand{\zh}[1]{\begin{CJK}{UTF8}{gbsn}\hspace{-15pt}\parbox{1.6cm}{#1}\hspace{-15pt}\end{CJK}}
\newcommand{\uphalfleftarrow}[1]{\overleftarrow{\raisebox{0pt}[.8\height]{#1}}}
\newcommand{\uphalfrightarrow}[1]{\overrightarrow{\raisebox{0pt}[.8\height]{#1}}}

The \ATT{} model characterizes the probability of an underlying punctuated tree $\utree$ given its corresponding unpunctuated tree $\dtree$, which is given by
\begin{align}
\p_\attp(\utree \mid \dtree) = \prod_{\node \in \dtree} \p_\attp(\lpuncteme_\node, \rpuncteme_\node \mid \node)\label{eq:underlying}
\end{align}
where $\lpuncteme_\node, \rpuncteme_\node \in \punctemes$ are the left and right \defn{punctemes} that $\utree$ attaches to the tree node $\node$.  Each puncteme \citep{krahn-2014} in the finite set $\punctemes$ is a string of 0 or more underlying punctuation \defn{tokens}.\footnote{Multi-token punctemes are occasionally useful.  For example, the puncteme \litf{...} might consist of either 1 or 3 tokens, depending on how the tokenizer works; similarly, the puncteme \litf{?!} might consist of 1 or 2 tokens. Also, if a single constituent of $\dtree$ gets surrounded by both parentheses and quotation marks, this gives rise to punctemes \litf{(``} and \litf{'')}.  (A better treatment would add the parentheses as a separate puncteme pair at a unary node above the quotation marks, but that would have required $\utree$ to introduce this extra node.)}
The probability $\p_\attp(\lpuncteme, \rpuncteme \mid \node)$ is given by a log-linear model

\noindent
\begin{align}\label{eq:loglin}
    \p_\attp(\lpuncteme, \rpuncteme | \node) 
  &\propto \begin{cases}
     \exp \attp^\top \vf(\lpuncteme, \rpuncteme, \node) & \!\!\!\text{if }(\lpuncteme,\rpuncteme)\in \punctemepairs_{\dep(\node)} \\
     0 & \!\!\!\text{otherwise}
   \end{cases}
\raisetag{10pt}
\end{align}

\noindent
where $\punctemes$ is the finite set of possible punctemes and $\punctemepairs_\dep \subseteq \punctemes^2$ gives the possible puncteme pairs for a node $\node$ that has dependency relation $\dep=\dep(\node)$ to its parent.  $\punctemes$ and $\punctemepairs_\dep$ are estimated heuristically from the tokenized surface data (\cref{sec:exp}).  $\vf(\lpuncteme, \rpuncteme, \node)$ is a sparse binary feature vector, and $\attp$ is the corresponding parameter vector of feature weights.  The feature templates in \appref{sec:template}%
\footnote{\label{ft:supp} The appendices are included only in this arXiv version, not in the TACL journal.}
  consider the symmetry between $l$ and $r$, and their compatibility with (a) the POS tag of $\node$'s head word, (b) the dependency paths connecting $\node$ to its children and the root of $\dtree$, (c) the POS tags of the words flanking the slots containing $l$ and $r$, (d) surface punctuation already added to $\node$'s subconstituents\cutforspace{ (bottom-up)}.

\subsection{From Underlying to Surface}\label{sec:nc}
\newcommand{\pnt}{s}
\newcommand{\tn}{t}
\newcommand{\pS}{S}
\newcommand{\pw}{h}
\newcommand{\rngT}{\mathcal{R}(\dtree)}
\newcommand{\nc}{\p(\cdot\mid i, \snt', \atr')}
\newcommand{\circled}[1]{\tikz[baseline=-.6ex]{\node[circle,draw,minimum size=4mm,inner sep=0pt] (char) {\small $#1$};}}
\newcommand{\dcircled}[1]{\tikz[baseline=-.6ex]{\node[circle,double,draw,minimum size=4mm,inner sep=0pt] (char) {\small $#1$};}}
\newcommand{\isym}{0}
\newcommand{\fsym}{\$}
\newcommand{\is}{\circled{$\isym$}}
\newcommand{\fs}{\dcircled{$\fsym$}}
\newcommand{\fstisym}{\text{$\wedge$}}
\newcommand{\fstfsym}{\$}
\newcommand{\fstis}{\circled{$\fstisym$}}
\newcommand{\fstfs}{\dcircled{$\fstfsym$}}
\newcommand{\pv}{\Sigma}
\newcommand{\tb}{\mathcal{\dtree}}
\newcommand{\reg}{\operatorname{R}}
\newcommand{\xdownarrow}[1]{%
  {\left\downarrow\vbox to #1{}\right.\kern-\nulldelimiterspace}
}

\setlength\tabcolsep{2pt}
\begin{table}[]

\centering
\begin{tabular}{ll|ll}
\hline
\multicolumn{2}{l|}{1. Point Absorption } & \multicolumn{2}{l}{3. Period Absorption} \\   
$\litf{\small ,,}\myrarw\litf{\small ,}$ \hspace{7pt}$\litf{\small ,.}\myrarw\litf{\small .}$ \hspace{7pt}$\litf{\small -,}\myrarw\litf{\small -}$&  & $\litf{\small .?}\myrarw \litf{\small ?}$ \hspace{7pt} $\litf{\small .!}\myrarw\litf{\small !}$  &  \\
   $\litf{\small -;}\myrarw\litf{\small ;}$ \hspace{7pt}$\litf{\small ;.}\myrarw\litf{\small .}$ & & $\litf{\small abbv.}\myrarw\litf{\small abbv}$ 
\vspace{-1.5pt}
\end{tabular}
\begin{tabular}{ll|ll}
\hline
 \multicolumn{2}{l|}{2. Quote Transposition} & \multicolumn{2}{l}{4. Bracket Absorptions}\\ 
 $\litf{\small '',}\myrarw\litf{\small ,''}$ \hspace{10pt}$\litf{\small ''.}\myrarw\litf{\small .''}$ &&  $\litf{\small ,)}\myrarw\litf{\small )}$\hspace{7pt}$\litf{\small -)}\myrarw\litf{\small )}$  &  $\litf{\small (,}\myrarw\litf{\small (}$\\ 
 & &  $\litf{\small ,''}\myrarw\litf{\small ''}$\hspace{10pt}$\litf{\small ``,}\myrarw\litf{\small ``}$&
\end{tabular}
\caption{\label{table:nunberg}Some of Nunberg's punctuation interaction rules in English, in priority order.  The absorption rules ensure that when there are two adjacent tokens, the ``weaker'' one is deleted (where the strength ordering is $\{\litf{?},\litf{!},\litf{(},\litf{)},\litf{``},\litf{''}\} > \litf{.} > \{\litf{;},\litf{:}\} > \litf{--} > \litf{,}$), except that bracketing tokens such as \litf{()} and \litf{``''} do not absorb tokens {\em outside} the material they bracket.}
\end{table}

From the tree $\utree$, we can read off the sequence of underlying punctuation tokens $\uslot_i$ at each slot $i$ between words.  Namely, $\uslot_i$ concatenates the right punctemes of all constituents ending at $i$ with the left punctemes of all constituents starting at $i$ (as illustrated by the examples in \cref{sec:intro} and \cref{fig:gen_story}). 
 The \NC{} model then transduces $\uslot_i$ to a surface token sequence $\sslot_i$, for each $i = 0,\ldots,n$ independently (where $n$ is the sentence length).

\paragraph{Nunberg's formalism}
Much like \citeposs{ChomskyHalle68} phonological grammar of English, \citeposs{nunberg1990linguistics} descriptive English punctuation grammar (\cref{table:nunberg}) can be viewed computationally as a priority string rewriting system, or \defn{Markov algorithm} \cite{markov-1960,caracciolo-1968}.
The system begins with a token string $\uslot$.  At each step it selects the highest-priority local rewrite rule that can apply, and applies it as far left as possible. When no more rules can apply, the final state of the string is returned as $\sslot$.

\paragraph{Simplifying the formalism} Markov algorithms are Turing complete.  Fortunately, \citet{johnson-1972} noted that in practice, phonological $\uslot \mapsto \sslot$ maps described in this formalism can usually be implemented with finite-state transducers (FSTs).  

For computational simplicity, we will formulate our punctuation model as a probabilistic FST (PFST)---a locally normalized left-to-right rewrite model \cite{cotterell14}.  The probabilities for each language must be learned, using gradient descent.  Normally we expect most probabilities to be near 0 or 1, making the PFST nearly deterministic (i.e., close to a subsequential FST).  However, permitting low-probability choices remains useful to account for typographical errors, dialectal differences, and free variation in the training corpus.

Our PFST \emph{generates} a surface string, but the invertibility of FSTs will allow us to work backwards when \emph{analyzing} a surface string (\cref{sec:inference}).

\newlength\myindent
\setlength\myindent{2em}
\newcommand\bindent{%
  \begingroup
  \setlength{\itemindent}{\myindent}
  \addtolength{\algorithmicindent}{\myindent}
}
\newcommand\eindent{\endgroup}

\newcommand{\eoi}{\$}
\newcommand{\udl}[1]{{\color{undb}\litf{#1}}}
\newcommand{\srf}[1]{{\color{undr}\litf{#1}}}
\newcommand{\sld}[1]{\underline{\litf{#1}}}
\newcommand{\comment}[1]{\hfill{\(\triangleright\) {\textit{#1}}}}
\newcommand{\commentt}[1]{\(\triangleright\) {\textit{#1}}}
\begin{figure}
\centering
\begin{tikzpicture}
\tikzstyle{every node}=[font=\scriptsize]
\node[draw=none] (a) at (0,0) {\sld{ab}\udl{cde}}; \node[text width=2cm] (aa) at ($(a) + (3,0)$) {\commentt{$\litf{ab}\;\myrarw\;\litf{ab}$}};
\node[draw=none] (b) at ($(a)+(0,-.3)$) {\srf{a}\sld{bc}\udl{de}};\node[draw=none,text width=2cm] (bb) at ($(aa) + (0,-.3)$) {\commentt{$\litf{bc}\;\myrarw\;\litf{b}$}};
\node[draw=none] (c) at ($(b)+(0,-.3)$) {\srf{a }\sld{bd}\udl{e}};\node[draw=none,text width=2cm] (cc) at ($(bb) + (0,-.3)$) {\commentt{$\litf{bd}\;\myrarw\;\litf{db}$}};
\node[draw=none] (d) at ($(c)+(0,-.3)$) {\srf{a d}\sld{be}};\node[draw=none,text width=2cm] (dd) at ($(cc) + (0,-.3)$) {\commentt{$\litf{be}\;\myrarw\;\litf{e}$}};
\node[draw=none] (f) at ($(d)+(0,-.3)$) {\srf{a d e}};
\end{tikzpicture}
\caption{\label{fig:slide} Editing $\litf{abcde}\mapsto\litf{ade}$ with a sliding window. (When an absorption rule maps 2 tokens to 1, our diagram leaves blank space that is not part of the output string.) At each step, the left-to-right process has already committed to the {\color{undr} green} tokens as output; has not yet looked at the {\color{undb} blue} input tokens; and is currently considering how to (further) rewrite the \underline{black} tokens. The right column shows the chosen edit.}
\end{figure}

\paragraph{A sliding-window model}\label{sec:sliding}
\cutforspace{Our specific model can be understood as follows.}
Instead of having rule priorities, we apply Nunberg-style rules within a 2-token window that slides over $\uslot$ in a single left-to-right pass (\cref{fig:slide}).
Conditioned on the current window contents $ab$, a single edit is selected stochastically: either $ab\myrarw ab$ (no change), $ab\myrarw b$ (left absorption), $ab\myrarw a$ (right absorption), or $ab\myrarw ba$ (transposition).  Then the window slides rightward to cover the next input token, together with the token that is (now) to its left.  $a$ and $b$ are always real tokens, never boundary symbols.  
$\ncp$ specifies the conditional edit probabilities.\footnote{\label{sec:strength}Rather than learn a separate edit probability distribution for each bigram $ab$, one could share parameters across bigrams.  For example, \cref{table:nunberg}'s caption says that ``stronger'' tokens tend to absorb ``weaker'' ones.  A model that incorporated this insight would not have to learn $O(|\pv|^2)$ separate absorption probabilities (two per bigram $ab$), but only $O(|\pv|)$ strengths (one per unigram $a$, which may be regarded as a 1-dimensional embedding of the punctuation token $a$).  We figured that the punctuation vocabulary $\pv$ was small enough (\cref{tb:data}) that we could manage without the additional complexity of embeddings or other featurization, although this does presumably hurt our generalization to rare bigrams.}

These specific edit rules (like Nunberg's) cannot insert new symbols, nor can they delete \emph{all} of the underlying symbols.  Thus, surface $\sslot_i$ is a good clue to $\uslot_i$: all of its tokens must appear underlyingly, and if $\sslot_i = \epsilon$ (the empty string) then $\uslot_i = \epsilon$.  

The model can be directly implemented as a PFST (\appref{sec:pfst}\footnoteref{ft:supp}) using \citeposs{cotterell14} more general PFST construction.%

Our single-pass formalism is less expressive than Nunberg's.  It greedily makes decisions based on at most one token of right context (``label bias'').  It cannot rewrite $\litf{'{}''.}\myrarw\litf{.'{}''}$ or $\litf{'',.}\myrarw\litf{.''}$ because the \litf{.} is encountered too late to percolate leftward; luckily, though, we can handle such English examples by sliding the window right-to-left instead of left-to-right.  We treat the sliding direction as a language-specific parameter.\footnote{We could have handled all languages uniformly by making $\geq 2$ passes of the sliding window (via a composition of $\geq 2$ PFSTs), with at least one pass in each direction.}

\newcommand{\pfst}{F}
\newcommand{\tra}[1]{\ra{#1}}

\subsection{Training Objective}
\label{sec:obj}
\newcommand{\obj}{\mathcal{J}}
\newcommand{\llk}{\mathcal{L}}
\newcommand{\pr}{\mathcal{C}}
\newcommand{\E}[2][]{\mathop{\mathbb{E}}_{{#1}}[#2]}

Building on \cref{eqn:conditmodel}, we train $\attp,\ncp$ to locally maximize the regularized conditional log-likelihood
\begin{align}\label{eqn:obj}
\hspace{-6pt}\Big( \sum_{\spunc,\dtree} \log p(\spunc \mid \dtree) 
      - \xi\cdot\E[\utree]{c(\utree)}^2 \Big)
     - \varsigma\cdot||\attp||^2 \raisetag{8pt}
\end{align}
\vspace{-17pt}

\noindent
where the sum is over a training treebank.\footnote{In retrospect, there was no good reason to square the $\E[\utree]{c(\utree)}$ term.  However, when we started redoing the experiments, we found the results essentially unchanged.}

The expectation $\E{\cdots}$ is over $\utree \sim \p(\cdot\mid \dtree,\spunc)$.  This \emph{generalized expectation} term provides \emph{posterior regularization} \cite{mann-mccallum-2010,ganchev2010posterior}, by encouraging parameters that reconstruct trees $\utree$ that use symmetric punctuation marks in a ``typical'' way. The function $c(\utree)$ counts the nodes in $\utree$ whose punctemes contain ``unmatched'' symmetric punctuation tokens: for example, \undb{)} is ``matched'' only when it appears in a right puncteme with \undb{(} at the comparable position in the same constituent's left puncteme. 
The precise definition\cutforspace{ uses a ``universal'' list of 12 symmetric pairs (\cref{sec:pr}\footnoteref{ft:supp})}
is given in \appref{sec:pr}.\footnoteref{ft:supp}

In our development experiments on English, the posterior regularization term was necessary to discover an aesthetically appealing theory of underlying punctuation.  When we dropped this term ($\xi = 0$) and simply maximized the ordinary regularized likelihood, we found that the optimization problem was underconstrained: different training runs would arrive at different, rather arbitrary underlying punctemes.  For example, one training run learned an \ATT{} model that used underlying \texttt{``.} to terminate sentences, along with a \NC{} model that absorbed the left quotation mark into the period.  By encouraging the underlying punctuation to be symmetric, we broke the ties.  We also tried making this a hard constraint ($\xi = \infty$), but then the model was unable to explain some of the training sentences at all, giving them probability of 0.  For example, \texttt{I went to the `` special place ''} cannot be explained, because \texttt{special place} is not a constituent.\footnote{Recall that the \NC{} model family (\cref{sec:nc}) requires the surface \texttt{``} before \texttt{special} to appear underlyingly, and also requires the surface $\epsilon$ after \texttt{special} to be empty underlyingly.  These hard constraints clash with the $\xi=\infty$ hard constraint that the punctuation around \texttt{special} must be balanced.  The surface \texttt{''} after \texttt{place} causes a similar problem: no edge can generate the matching underlying \texttt{``}.}

\newcommand{\word}{\operatorname{word}}
\newcommand{\nl}{l}
\newcommand{\child}{\operatorname{chr}}
\newcommand{\parent}{\operatorname{par}}
\newcommand{\rt}{\depf{root}}
\newcommand{\df}{\triangleq}
\section{Inference}\label{sec:inference}

\newcommand{\fpe}{\rho}
\newcommand{\Fpe}{B}
\newcommand{\fp}{\vec{\fpe}}
\newcommand{\Fp}{\vec{\Fpe}}
\newcommand{\Real}{\mathbb{R}}
\newcommand{\Mfsa}{\vec{M}}
\newcommand{\Fu}{\vec{U}}
\newcommand{\smr}{\vec{R}}
\newcommand{\smrw}{M}
\newcommand{\smrW}{\vec{\smrw}}
\newcommand{\sdtr}{\tau}
\newcommand{\sutr}{\sdtr'}
\newcommand{\splus}{\oplus}
\newcommand{\sSum}{\bigoplus}
\newcommand{\stimes}{\otimes}
\newcommand{\sProd}{\bigotimes}
\newcommand{\pfstpc}[1]{\pfst\circ#1}
\newcommand{\wfst}{\pfstpc{\sslot_i}}
\newcommand{\str}{\tau}
\newcommand{\nwfst}{N}
\newcommand{\paths}{\mathcal{P}}
\newcommand{\stt}[1]{\tc{#1}}
\newcommand{\pth}{\rho}
\newcommand{\vpth}{\vec{\pth}}

In principle, working with the model \eqref{eqn:model} is straightforward, thanks to the closure properties of formal languages.  Provided that $\psyn$ can be encoded as a weighted CFG, it can be composed with the weighted tree transducer $p_\attp$ and the weighted FST $p_\ncp$ to yield a new weighted CFG \citep[similarly to][]{BarHillel61,nederhof2003probabilistic}.
  Under this new grammar, one can recover the optimal $T,T'$ for $\ssent$ by dynamic programming, or sum over $T,T'$ by the inside algorithm to get the likelihood $p(\ssent)$.
A similar approach
was used by \citet{D08-1025} with a different FST noisy channel.

In this paper we assume that $\dtree$ is observed, allowing us to work with \cref{eqn:conditmodel}.  This cuts the computation time from $O(n^3)$ to $O(n)$.\footnote{We do $O(n)$ multiplications of $N \times N$ matrices where $N = O(\text{\# of punc types}\cdot\text{max \# of punc tokens per slot})$.}
Whereas the inside algorithm for \eqref{eqn:model} must consider $O(n^2)$ possible constituents of $\ssent$ and $O(n)$ ways of building each, our algorithm for \eqref{eqn:conditmodel} only needs to iterate over the $O(n)$ true constituents of $\dtree$ and the 1 true way of building each.  However, it must still consider the $|\punctemepairs_\dep|$ puncteme pairs for each constituent.

\newcommand{\rightcomment}[1]{\(\triangleright\) {\footnotesize\textit{#1}}}
\algrenewcommand{\algorithmiccomment}[1]{\hfill \rightcomment{#1}}
\algnewcommand{\LineComment}[1]{\State \rightcomment{#1}}
\algnewcommand{\LinesComment}[1]{\State \rightcomment{\parbox[t]{\linewidth-\leftmargin-\widthof{\(\triangleright\) }}{#1}}\smallskip}
\newcommand{\la}[1]{\hspace{-3pt}\xxleftarrow{1}{#1}\hspace{-3pt}}
\definecolor{ec}{RGB}{0,0,255}
\newcommand{\eccommand}[1]{{\color{ec}#1}}
\newcommand{\inside}{In}
\begin{algorithm}
  \caption{\label{alg:inside} The algorithm for scoring a given $(\dtree, \spunc)$ pair.  
  The code in \textcolor{ec}{blue} is used during training to get the posterior regularization term in \eqref{eqn:obj}.}
  \begin{algorithmic}[1]
    \INPUT $\dtree$, $\spunc$\Comment{Training pair (omits $\utree,\upunc$)}
    \OUTPUT $\p(\spunc\mid\dtree)$, \textcolor{ec}{$\E{c(\utree)}$}%
    \Procedure{TotalScore}{$\dtree$, $\spunc$}
    \For {$i = 1 \To n$}
    \State compute WFSA $(\Mfsa_i, \vec{\lambda}_i, \vec{\rho}_i)$\label{alg:line:fst}
    \EndFor
    \State \textcolor{ec}{$E \gets 0$ } \Comment{exp.\@ count of unmatched punctemes} \label{alg:line:unpair}
    \Procedure{\inside}{$\node$} \Comment{$\node \in T$}
    \State $i,k\gets{}$slots at left, right of $\node$ constit
    \State $j\gets{}$slot at right of $\node$ headword
    \State $\!\smrW_{\text{left}} \!\gets\! (\prod_{\node' \in \mathrm{leftkids}(\node)} \Call{\inside}{\node'}) \vec{\rho}_{j-1}$\label{alg:line:mult1}
    \State $\!\smrW_{\text{right}} \!\gets\! \vec{\lambda}_j^\top (\prod_{\node' \in \mathrm{rightkids}(\node)} \Call{\inside}{\node'})$\label{alg:line:mult2}
    \State $\smrW' \gets \smrW_{\text{left}}\cdot 1 \cdot \smrW_{\text{right}}$\label{alg:line:mult3}\label{alg:line:one} \Comment{$\Real^{N_j \times 1}, \Real^{1 \times N_j}$}
    \State $\smrW \gets \vec{0}$ \Comment{$\Real^{N_i\times N_k}$}\label{alg:line:zero}
    \For {$(\lpuncteme, \rpuncteme) \in \punctemepairs_{\dep(\node)}$} \label{alg:line:lr}
        \State $p\gets \p_\attp(\lpuncteme, \rpuncteme \mid \node)$ \label{alg:line:p}
        \State $\smrW \gets \smrW + p \cdot \Mfsa_i(\lpuncteme)\vec{M}'\Mfsa_k(\rpuncteme)$\label{alg:line:plus}\label{alg:line:mult4}
        \State \textcolor{ec}{$E \gets E + p \cdot \mathbbm{1}_{\lpuncteme,\rpuncteme~\text{have unmatched punc}}$} 
    \EndFor
    \State \Return $\smrW$ \Comment{$\Real^{N_i\times N_k}$}
    \EndProcedure
    \State $M_{\text{root}} \gets \Call{\inside}{\mathrm{root}(\dtree)}$
    \State \Return $\vec{\lambda}_0^\top M_{\text{root}} \fp_{n}$\eccommand{, $E$} \Comment{$\Real\eccommand{, \Real}$}
    \label{alg:line:mult5} 
  \EndProcedure
\end{algorithmic}
\end{algorithm}

\subsection{Algorithms} \label{sec:algo}

\newcommand{\tightudl}[1]{\underline{#1}}
\newcommand{\detclr}{undr!50}
\newcommand{\detcircled}[1]{\tikz[baseline=-.6ex]{\node[circle,draw,minimum size=4mm,inner sep=0pt,fill=\detclr] (char) {\small \tightudl{$#1$}};}}
\newcommand{\detdcircled}[1]{\tikz[baseline=-.6ex]{\node[circle,double,draw,minimum size=4mm,inner sep=0pt,fill=\detclr] (char) {\small \tightudl{$#1$}};}}
\newcommand{\tddc}[1]{\tikz[baseline=-.6ex]{\node[circle,double,draw,minimum size=4mm,inner sep=0pt,fill=\detclr] (char) {\tiny \tightudl{$#1$}};}}
\newcommand{\tc}[1]{\tikz[baseline=-.6ex]{\node[circle,draw,minimum size=4mm,inner sep=0pt] (char) {\scriptsize $#1$};}}
\newcommand{\ttc}[1]{\tikz[baseline=-.6ex]{\node[circle,draw,minimum size=4mm,inner sep=0pt,text width={width("y")}] (char) {\tiny $#1$};}}
\newcommand{\detisym}{\tightudl{0}}
\newcommand{\detis}{\detcircled{$\isym$}}
\newcommand{\ttra}[1]{\hspace{-1pt}\ra{#1}\hspace{-1pt}}

Given an input sentence $\ssent$ of length $n$, our job is to sum over possible trees $\utree$ that are consistent with $\dtree$ and $\ssent$, or to find the best such $\utree$.  This is roughly a lattice parsing problem---made easier by knowing $\dtree$.  However, the possible $\usent$ values are characterized not by a lattice but by a \emph{cyclic} WFSA (as $|\uslot_i|$ is unbounded whenever $|\sslot_i|>0$). 

For each slot $0 \leq i \leq n$, transduce the surface punctuation string $\sslot_i$ by the \emph{inverted} PFST for $p_\ncp$ to obtain a weighted finite-state automaton (WFSA) that describes \emph{all possible} underlying strings $\uslot_i$.\footnote{\label{fn:wfsa}Constructively, compose the $\uslot$-to-$\sslot$ PFST (from the end of \cref{sec:nc}) with a straight-line FSA accepting only $\sslot_i$, and project the resulting WFST to its input tape \cite{pereira-riley-1996}, as explained at the end of \appref{sec:pfst}.}
This WFSA accepts each possible $\uslot_i$ with weight $\p_\ncp(\sslot_i \mid \uslot_i)$.  If it has $N_i$ states, we can represent it \cite{berstel-reutenauer-1988} with a family of sparse weight matrices $\Mfsa_i(\utok) \in \Real^{N_i \times N_i}$, whose element at row $s$ and column $t$ is the weight of the $s \to t$ arc labeled with $\utok$, or 0 if there is no such arc.  Additional vectors $\vec{\lambda}_i, \vec{\rho}_i \in \Real^{N_i}$ specify the initial and final weights.
 ($\vec{\lambda}_i$ is one-hot if the PFST has a single initial state, of weight 1.)

For any puncteme $\lpuncteme$ (or $r$) in $\punctemes$, we define $\Mfsa_i(\lpuncteme) = \Mfsa_i(\lpuncteme_1)\Mfsa_i(\lpuncteme_2)\cdots\Mfsa_i(\lpuncteme_{|\lpuncteme|})$, a product over the 0 or more tokens in $\lpuncteme$.  This gives the total weight of all $s \to^* t$ WFSA paths labeled with $\lpuncteme$.

The subprocedure in \cref{alg:inside} essentially extends this to obtain a new matrix $\Call{\inside}{\node} \in \Real^{N_i \times N_k}$, where the subtree rooted at $\node$ stretches from slot $i$ to slot $k$.  Its element $\Call{\inside}{\node}_{st}$ gives the total weight of all \defn{extended paths} in the $\usent$ WFSA from state $s$ at slot $i$ to state $t$ at slot $k$.  An extended path is defined by a choice of underlying punctemes at $\node$ and all its descendants.  These punctemes determine an $s$-to-final path at $i$, then initial-to-final paths at $i+1$ through $k-1$, then an initial-to-$t$ path at $k$.  The weight of the extended path is the product of all the WFSA weights on these paths (which correspond to transition probabilities in $p_\ncp$ PFST) times the probability of the choice of punctemes (from $p_\attp$).

This inside algorithm computes quantities 
needed for training (\cref{sec:obj}).  Useful variants arise via well-known methods for weighted derivation forests \cite{berstel-reutenauer-1988,goodman-1999,li-eisner-2009,eisner-2016}.  

Specifically, to modify \cref{alg:inside} to \emph{maximize} over $\utree$ values (\crefrange{sec:corr}{sec:gd}) instead of summing over them, we switch to the \defn{derivation semiring} \cite{goodman-1999}, as follows. Whereas $\Call{\inside}{\node}_{st}$ used to store the \emph{total} weight of all extended paths from state $s$ at slot $i$ to state $t$ at slot $j$, now it will store the weight of the \emph{best} such extended path.  It will also store that extended path's
  choice of underlying punctemes, in the form of a puncteme-annotated version of the subtree of $\dtree$ that is rooted at $\node$.  This is a potential subtree of $\utree$. 

Thus, each element of $\Call{\inside}{\node}$ has the form $(r,\struct)$ where $r \in \Real$ and $\struct$ is a tree.  We define addition and multiplication over such pairs:
\begin{align}
(r,\struct) + (r',\struct') &= 
    \begin{cases}
      (r,\struct) & \text{if } r > r' \\
      (r',\struct') & \text{otherwise}
    \end{cases} \label{eqn:oplus} \\
(r,\struct) \cdot (r',\struct') &= (rr', \struct\struct') \label{eqn:otimes}
\end{align}
where $\struct\struct'$ denotes an ordered combination of two trees.  Matrix products $\vec{U}\vec{V}$ and scalar-matrix products $p\cdot\vec{V}$ are defined in terms of element addition and multiplication as usual:

\noindent
\begin{align}
  (\vec{U}\vec{V})_{st} &= \textstyle{\sum_r} \vec{U}_{sr} \cdot \vec{V}_{rt} \\
  (p\cdot \vec{V})_{st} &= p \cdot \vec{V}_{st} 
\end{align}

What is $DD'$? For presentational purposes, it is convenient to represent a punctuated dependency tree as a bracketed string.  For example, the underlying tree $\utree$ in \cref{fig:gen_story} would be {\small 
\texttt{[~[``~Dale~''] means [``~[~river~] valley~'']~]}}
where the words correspond to nodes of $\dtree$.
In this case, we can represent every $\struct$ as a \emph{partial} bracketed string and define $\struct\struct'$ by string concatenation.  This presentation ensures that multiplication \eqref{eqn:otimes} is a complete and associative (though not commutative) operation, as in any semiring.  As base cases, each real-valued element of $\smrW_i(\lpuncteme)$ or $\smrW_k(\rpuncteme)$ is now paired with the string \texttt{[$\lpuncteme$} or \texttt{$\rpuncteme$]} respectively,%
\footnote{We still construct the real matrix $\smrW_i(\lpuncteme)$ by \emph{ordinary} matrix multiplication before pairing its elements with strings.  This involves summation of real numbers: each element of the resulting real matrix is a marginal probability, which sums over possible PFST paths (edit sequences) that could map the underlying puncteme $\lpuncteme$ to a certain substring of the surface slot $\sslot_i$.  Similarly for $\smrW_k(\rpuncteme)$.}
and the real number 1 at \cref{alg:line:one} is paired with the string $\node$.  The real-valued elements of the $\vec{\lambda}_i$ and $\vec{\rho}_i$ vectors and the $\vec{0}$ matrix at \cref{alg:line:zero} are paired with the empty string $\epsilon$, as is the real number $p$ at \cref{alg:line:p}.

In practice, the $\struct$ strings that appear within the matrix $\smrW$ of \cref{alg:inside} will always represent complete punctuated trees.  Thus, they can actually be represented in memory as such, and different trees may share subtrees for efficiency (using pointers).  The product in \cref{alg:line:one} constructs a matrix of trees with root $w$ and differing sequences of left/right children, while the product in \cref{alg:line:plus} annotates those trees with punctemes $\lpuncteme,\rpuncteme$.

To sample a possible $\utree$ from the derivation forest in proportion to its probability (\cref{sec:restore}),
we use the same algorithm but replace \cref{eqn:oplus} with
\begin{align}
  (r,\struct) + (r',\struct') &= 
  \begin{cases}
    (r+r',\struct) & \text{if } u < \frac{r}{r+r'} \\
    (r+r',\struct') & \text{otherwise} 
  \end{cases} \nonumber
\end{align}
with $u \sim \text{Uniform}(0,1)$ being a random number.

\subsection{Optimization}
Having computed the objective \eqref{eqn:obj}, we find the gradient via automatic differentiation, and optimize $\attp,\ncp$ via Adam \cite{kingma2014adam}---a variant of stochastic gradient decent---with learning rate 0.07, batchsize 5, sentence per epoch 400, and L2 regularization.  (These hyperparameters, along
with the regularization coefficients $\varsigma$ and $\xi$ from \cref{eqn:obj}, were tuned on dev data (\cref{sec:exp}) for each language respectively.)
We train the punctuation model for 30 epochs. The initial \NC{} parameters ($\ncp$) are drawn from ${\cal N}(0,1)$, and the initial \ATT{} parameters ($\attp$) are drawn from ${\cal N}(0,1)$ (with one minor exception described in \appref{sec:template}).

\section{Intrinsic Evaluation of the Model}
\newcommand{\tbf}[1]{\textsf{#1}}
\label{sec:exp}\label{sec:perplexity}\label{sec:data}\label{sec:baseline}

\paragraph{Data.} Throughout \crefrange{sec:exp}{sec:app},
we will examine the punctuation model on a subset of the Universal Dependencies (UD) version 1.4 \cite{UNIVDEP-1.4}\cutforspace{\footnote{It contains the latest treebank of English as a Second Language (\tbf{en\_esl}) data, which is used in our experiments.}}---a collection of dependency treebanks across 47 languages with unified POS-tag and dependency label sets.  Each treebank has designated training, development, and test portions.  We experiment on Arabic, English, Chinese, Hindi, and Spanish (\cref{tb:data})---languages with diverse punctuation vocabularies and punctuation interaction rules, not to mention script directionality.  
For each treebank, we use the tokenization provided by UD, and take the punctuation tokens (which may be multi-character, such as \litf{...})~to be the tokens with the \posf{PUNCT} tag.  
We replace each straight double quotation mark \litf{"} with either \litf{``} or \litf{''} as appropriate, and similarly for single quotation marks.\footnote{For \tbf{en} and \tbf{en\_esl}, \litf{``} and \litf{''} are distinguished by language-specific part-of-speech tags. For the other 4 languages, we identify two \litf{"} dependents of the same head word, replacing the left one with  \litf{``} and the right one with \litf{''}.  }
We split each non-punctuation token that ends in \litf{.} (such as \litf{etc.}) into a shorter non-punctuation token (\litf{etc}) followed by a special punctuation token called the ``abbreviation dot'' (which is distinct from a period).
We prepend a special punctuation mark \litf{\bossym} to every sentence $\ssent$, which can serve to absorb an initial comma, for example.\footnote{For symmetry, we should also have added a final mark.}
We then replace each token with the special symbol \litf{UNK} if its type appeared fewer than $5$ times in the training portion.  This gives the surface sentences.

To estimate the vocabulary $\punctemes$ of \emph{underlying} punctemes, we simply collect all \emph{surface} token sequences $\sslot_i$ that appear at any slot in the training portion of the processed treebank.  This is a generous estimate.  Similarly, we estimate $\punctemepairs_\dep$ (\cref{sec:gen}) as all pairs $(\lpuncteme,\rpuncteme) \in \punctemes^2$ that flank any $\dep$ constituent.

Recall that our model generates surface punctuation given an unpunctuated dependency tree.  We train it on each of the 5 languages independently.  We evaluate on conditional perplexity, which will be low if the trained model successfully assigns a high probability to the actual surface punctuation in a held-out corpus of the same language.
\begin{table}
\setlength\tabcolsep{2pt}
\centering
\resizebox{0.47\textwidth}{!}{
\begin{tabular}{l|c|c|c|c|c}
    Language             &Treebank    & \#Token &   \%Punct & \#Omit & \#Type \\\hline\hline
    Arabic               &\small\tbf{ar}  &    282K    &    7.9   & 255  & 18      \\\hline
    Chinese              &\small\tbf{zh}  &    123K    &    13.8  & 3  &  23    \\\hline
\multirow{2}{*}{English} &\small\tbf{en}  &    255K    &    11.7  & 40  &  35    \\
                         &\small\tbf{en\_esl}  &    97.7K    &   9.8   & 2  &  16    \\\hline
    Hindi                &\small\tbf{hi}  &    352K    &    6.7   & 21    &  15  \\\hline
    Spanish              &\small\tbf{es\_ancora}  &    560K    &   11.7  & 25   &  16   \\
\end{tabular}}
\vspace{-2pt}
\caption{\label{tb:data}Statistics of our datasets. ``Treebank'' is the UD treebank identifier, ``\#Token'' is the number of tokens, ``\%Punct'' is the percentage of punctuation tokens, ``\#Omit'' is the small number of sentences containing non-leaf punctuation tokens (see \cref{fn:omit}), and ``\#Type'' is the number of punctuation types after preprocessing. (Recall from \cref{sec:data} that preprocessing distinguishes between left and right quotation mark types, and between abbreviation dot and period dot types.)}
\end{table}

\paragraph{Baselines.} 
We compare our model against three baselines to show that its complexity is necessary.
Our first baseline is an ablation study that does not use latent underlying punctuation, but generates the surface punctuation directly from the tree.  (To implement this, we fix the parameters of the noisy channel so that the surface punctuation equals the underlying with probability 1.)  If our full model performs significantly better, it will demonstrate the importance of a distinct underlying layer. 

Our other two baselines ignore the tree structure, so if our full model performs significantly better, it will demonstrate that conditioning on explicit syntactic structure is useful.  These baselines are based on previously published approaches that reduce the problem to tagging: 
\citet{bilstm-crf} use a BiLSTM-CRF tagger with bigram topology;
\citet{tilk2016bidirectional} use a BiGRU tagger with attention.  
In both approaches, the model is trained to tag each slot $i$ with the correct string $\sslot_i \in \tags$ (possibly $\epsilon$ or $\litf{\bossym}$).
These are discriminative probabilistic models (in contrast to our generative one).  Each gives a probability distribution over the taggings (conditioned on the unpunctuated sentence), so we can evaluate their perplexity.\footnote{These methods learn word embeddings that optimize conditional log-likelihood on the  punctuation restoration training data.  They might do better if these embeddings were shared with other tasks, as multi-task learning might lead them to discover syntactic categories of words.} 

\paragraph{Results.}

As shown in \cref{tab:perp}, our full model beats the baselines in perplexity in all 5 languages. Also,
in 4 of 5 languages, allowing a trained \NC{} (rather than the identity map) significantly improves the perplexity.

\definecolor{oracle}{rgb}{0.5, 0.5, 0.5}
\newcommand{\orcf}[1]{{\color{oracle}#1}}
\setlength\tabcolsep{3pt}
\begin{table}[]
\centering
\begin{tabular}{l|c|c|ccc}
    & Attn.  & CRF & \ATT{} & +\textsc{NC} & \textsc{Dir}\\ \hline
Arabic  &1.4676    &1.3016  & 1.2230 & \textbf{1.1526} & L\\
Chinese & 1.6850   &1.4436 &  1.1921 & \textbf{1.1464} & L\\
English &1.5737     &\textbf{1.5247} &  1.5636 & \textbf{1.4276} & R\\
Hindi   & 1.1201   & 1.1032&  1.0630 & \textbf{1.0598} & L\\
Spanish & 1.4397   &\textbf{1.3198}  & \textbf{1.2364}& \textbf{1.2103} & R \\
\end{tabular}
\caption{\label{tab:perp}Results of the conditional perplexity experiment (\cref{sec:perplexity}), reported as perplexity per punctuation slot, where an unpunctuated sentence of $n$ words has $n+1$ slots. Column ``Attn.\@'' is the BiGRU tagger with attention, and ``CRF'' stands for the BiLSTM-CRF tagger.  ``\ATT{}'' is the ablated version of our model where surface punctuation is directly attached to the nodes.  Our full model ``+\textsc{NC}'' adds \NC{} to transduce the attached punctuation into surface punctuation. \textsc{Dir} is the learned direction (\cref{sec:sliding}) of our full model's noisy channel PFST: \underline{L}eft-to-right or \underline{R}ight-to-left.  Our models are given oracle parse trees $\dtree$. The best perplexity is \textbf{boldfaced}, along with all results that are not significantly worse (paired permutation test, $p < 0.05$).}
\end{table}

\section{Analysis of the Learned Grammar}
\label{sec:qual}
\subsection{Rules Learned from the Noisy Channel}
\label{sec:rule}

\begin{figure}
\includegraphics[page=1,width=0.5\textwidth]{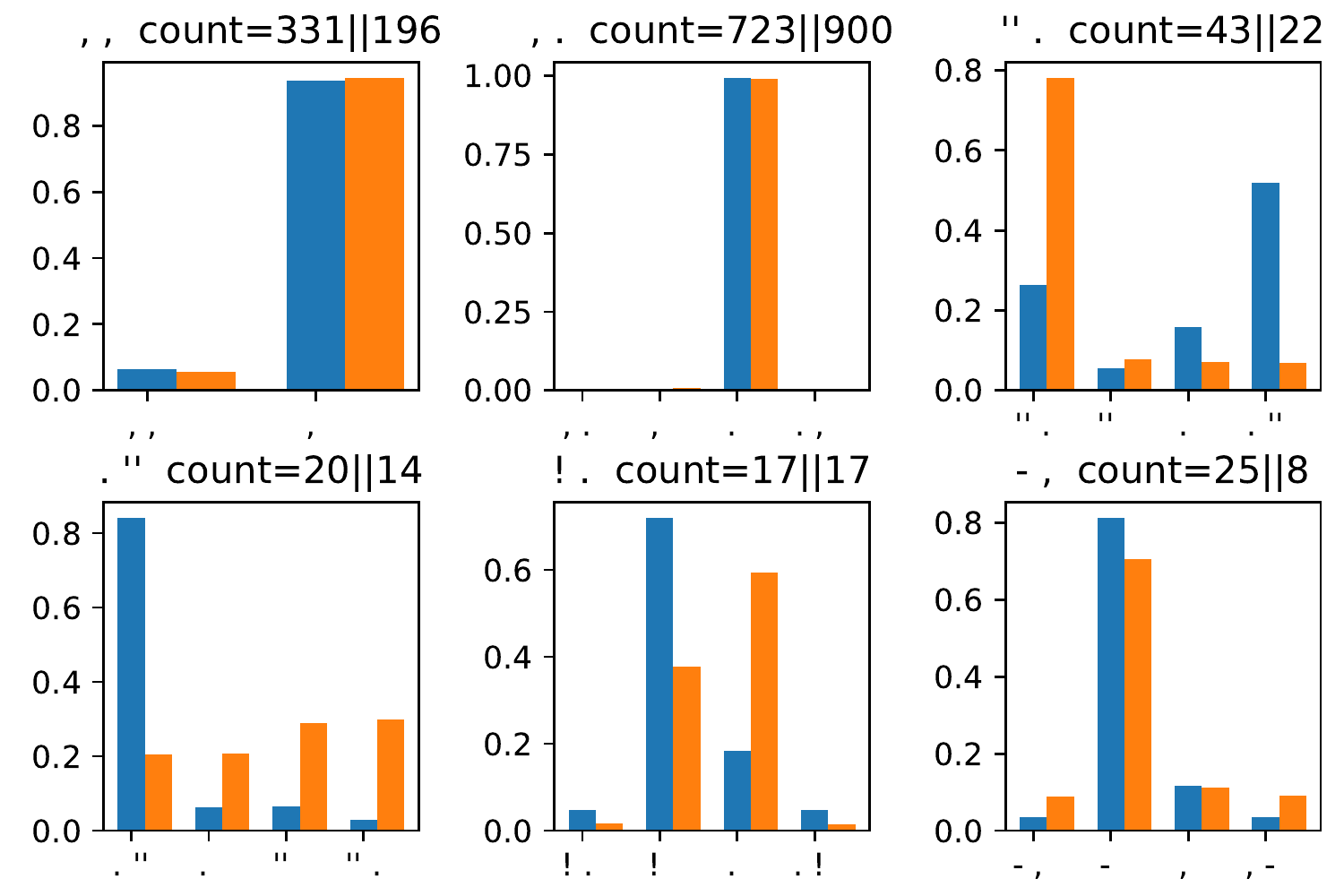}
\vspace{-20pt}
\caption{\label{fig:en-point}Rewrite probabilities learned for English, averaged over the last 4 epochs on \tbf{en} treebank (blue bars) or \tbf{en\_esl} treebank (orange bars). 
The header above each figure is the underlying punctuation string (input to \NC{}). 
The two counts in the figure headers are the number of occurrences of the underlying punctuation strings in the 1-best reconstruction of underlying punctuation sequences (by \cref{alg:inside}) respectively in the \tbf{en} and \tbf{en\_esl} treebank. Each bar represents one surface punctuation string (output of \NC{}), its height giving the probability. 
}
\end{figure}

We study our learned probability distribution over noisy channel rules ($ab\myrarw b$, $ab\myrarw a$, $ab\myrarw ab$, $ab\myrarw ba$) for English.
The probability distributions corresponding to six of Nunberg's English rules are shown in \cref{fig:en-point}. 
By comparing the orange and blue bars, observe that the model trained on the \tbf{en\_cesl} treebank learned different quotation rules from the one trained on the \tbf{en} treebank. This is because \tbf{en\_cesl} follows British style,\cutforspace{\footnote{The ESL data comes from Cambridge ESOL First Certificate in English (FCE) examination, which is a British exam, so the corrected punctuation follows the British style.}} whereas \tbf{en} has American-style quote transposition.\footnote{American style places commas and periods inside the quotation marks, even if they are not logically in the quote. British style (more sensibly) places unquoted periods and commas in their logical place, sometimes outside the quotation marks if they are not part of the quote.}

We now focus on the model learned from the \tbf{en} treebank.  Nunberg's rules are deterministic, and our noisy channel indeed learned low-entropy rules, in the sense that for an input $ab$ with underlying count $\geq 25$,\footnote{For rarer underlying pairs $ab$, the estimated distributions sometimes have higher entropy due to undertraining.} at least one of the possible outputs ($a$, $b$, $ab$ or $ba$) always has probability $> 0.75$.
The one exception is $\litf{''.}\myrarw\litf{.''}$ for which the argmax output has probability $\approx 0.5$,
because writers do not apply this quote transposition rule consistently. As shown by the blue bars in \cref{fig:en-point}, the high-probability transduction rules are consistent with Nunberg's hand-crafted deterministic grammar in \cref{table:nunberg}.

Our system has high precision when we look at the confident rules.  Of the 24 learned edits with conditional probability $> 0.75$, Nunberg lists 20.

Our system also has good recall.  Nunberg's hand-crafted schemata consider 16 punctuation types and generate a total of 192 edit rules, including the specimens in \cref{table:nunberg}.  That is, of the $16^2 = 256$ \emph{possible} underlying punctuation bigrams $ab$, $\frac{3}{4}$ are supposed to undergo absorption or transposition.  Our method achieves fairly high recall, in the sense that when Nunberg proposes $ab \myrarw \gamma$, our learned $p(\gamma \mid ab)$ usually ranks highly among all probabilities of the form $p(\gamma' \mid ab)$.  75 of Nunberg's rules got rank 1, 48 got rank 2,
and the remaining 69 got rank $> 2$.  The mean reciprocal rank was 0.621.  Recall is quite high when we restrict to those Nunberg rules $ab \myrarw \gamma$ for which our model is confident how to rewrite $ab$, in the sense that some $p(\gamma' \mid ab) > 0.5$.  (This tends to eliminate rare $ab$: see \cref{sec:strength}.)  Of these 55 Nunberg rules, 38 rules got rank 1, 15 got rank 2,
and only 2 got rank worse than 2. The mean reciprocal rank was 0.836. 

?`What about Spanish?  Spanish uses inverted question marks \litf{?`} and exclamation marks \litf{!`}, which form symmetric pairs with the regular question marks and exclamation marks. If we try to extrapolate to Spanish from Nunberg's English formalization, the English mark most analogous to  \litf{?`} is \litf{(}. Our learned noisy channel for Spanish (not graphed here) includes the high-probability rules $\litf{,?`}\myrarw\litf{,?`}$ and $\litf{:?`}\myrarw\litf{:?`}$ and $\litf{?`,}\myrarw\litf{?`}$ which match Nunberg's treatment of \litf{(} in English. 

\subsection{Attachment Model}
\label{sec:underly}

\newcommand{\kerry}{
\hspace{-5pt}\begin{dependency}[baseline=-1em,label style={above,draw=none,fill=none,inner sep=0ex,outer sep=.3ex},edge vertical padding=-7pt]
\tikzstyle{every node}=[font=\tiny]
\begin{deptext}[column sep=.1em, row sep=-.6em]
    \litf{\undb{\bossym,}}\&[-7pt]\litf{Earlier} \&[-7pt]\litf{\undb{,}}\&[-3pt] \litf{Kerry}\&[-3pt]\litf{said}\&[0pt]\litf{\undb{,``}}\&[-5pt]\litf{...}\&[-5pt]\litf{\undb{,}}\& [-7pt] \litf{in} \&[-3pt]\litf{fact} \&[-7pt]\litf{\undb{,}} \&[-4pt]\litf{answer} \&[-3pt]\litf{the} \&[-3pt]\litf{question\undb{''.}} \\
    \litf{\undr{\bossym}}\&\litf{Earlier}\&\litf{\undr{,}}\&\litf{Kerry}\&\litf{said}\&\litf{\undr{,``}}\&\litf{...}\&\litf{\undr{,}}\&\litf{in}\&\litf{fact}\&\litf{\undr{,}} \&\litf{answer}\&\litf{the}\&\litf{question\undr{.''}} \\
\end{deptext}
\deproot[edge unit distance=.5ex]{5}{\scriptsize\depf{root}\undb{.}}
\depedge[edge unit distance=.5ex]{5}{2}{\scriptsize\undb{,}\depf{advmod}\undb{,}}
\depedge[edge unit distance=.3ex]{5}{12}{\scriptsize\undb{,}\undb{``}\depf{ccomp}\undb{''}}
\depedge[edge unit distance=.2ex]{12}{10}{\scriptsize\undb{,}\depf{nmod}\undb{,}}
\depedge[edge unit distance=.4ex,edge horizontal padding=-.5pt]{10}{9}{}
\end{dependency}\hspace{-5pt}}
\newcommand{\oxford}{
\hspace{-.5em}\begin{dependency}[baseline=-.1em,label style={above,draw=none,fill=none,inner sep=0ex,outer sep=.3ex},edge vertical padding=-4pt]
\tikzstyle{every node}=[font=\scriptsize]
\begin{deptext}[column sep=.3em, row sep=-.6em]
    \litf{Section}\&\litf{1}\&[8pt]\&\litf{\undb{,}2}\&[-5pt]\litf{\undb{,}}\&[-4pt]\litf{\undb{,}...}\&[-7pt]\litf{7}\&[-7pt]\litf{\undb{,}}\&[-5pt]\litf{and}\&\litf{8}\&[-7pt]\litf{...} \\
    \litf{Section}\&\litf{1}\&\&\litf{\undr{,}2}\&\&\litf{\undr{,}...}\&\litf{7}\&\litf{\undr{,}}\&\litf{and}\&\litf{8}\&\litf{...}\\
\end{deptext}
\depedge[edge unit distance=.3ex]{2}{4}{\scriptsize\undb{,}\depf{conj}\undb{,}}
\depedge[edge unit distance=.4ex]{2}{7}{\scriptsize\undb{,}\depf{conj}\undb{,}}
\depedge[edge unit distance=.45ex]{2}{10}{\scriptsize\depf{conj}}
\depedge[edge unit distance=.6ex]{10}{9}{\scriptsize\depf{cc}}
\end{dependency}\hspace{-.5em}}

What does our model learn about how dependency relations are marked by underlying punctuation? 

\kerry 
The above example\footnote{\scriptsize [\tbf{en}] \texttt{Earlier, Kerry said, ``Just  because  you  get  an  honorable  discharge  does  not, in  fact, answer  that  question.''} } illustrates the use of specific puncteme pairs to set off the \depf{advmod}, \depf{ccomp}, and \depf{nmod} relations. Notice that \litf{said} takes a complement (\depf{ccomp}) that is symmetrically quoted but also left delimited by a comma, which is indeed how direct speech is punctuated in English. This example also illustrates quotation transposition. The top five relations that are most likely to generate symmetric punctemes and their top $(\lpuncteme,\rpuncteme)$ pairs are shown in \cref{tab:top}. 

\oxford{} \\
The above example\footnote{\scriptsize [\tbf{en}] \texttt{Sections 1, 2, 5, 6, 7, and 8  will  survive  any  termination  of  this  License.}}
shows how our model handles commas in conjunctions of 2 or more phrases.  UD format dictates that each conjunct after the first is attached by the \depf{conj} relation.  As shown above, each such conjunct is surrounded by underlying commas (via the \texttt{N}.\litf{,}.\litf{,}.\depf{conj} feature from \appref{sec:template}), except for the one that bears the conjunction \texttt{and} (via an even stronger weight on the \texttt{C}.$\epsilon$.$\epsilon$.$\uphalfrightarrow{\depf{conj}}$.\depf{cc} feature).  Our learned feature weights indeed yield $p(\ell=\epsilon,r=\epsilon) > 0.5$ for the final conjunct in this example.  Some writers omit the ``Oxford comma'' before the conjunction: this style can be achieved simply by changing ``surrounded'' to ``preceded'' (that is, changing the \texttt{N} feature to \texttt{N}.\litf{,}.$\epsilon$.\depf{conj}).

\section{Performance on Extrinsic Tasks}
\label{sec:app}

We evaluate the trained punctuation model by using it in the following three tasks. 

\subsection{Punctuation Restoration}
\label{sec:restore}
In this task, we are given a depunctuated sentence $\dsent$\footnote{\label{fn:omit} To depunctuate a treebank sentence, we remove all tokens with POS-tag \texttt{PUNCT} or dependency relation \texttt{punct}.  These are almost always leaves; else we omit the sentence.} and must restore its (surface) punctuation.  Our model supposes that the observed punctuated sentence $\ssent$ would have arisen via the generative process \eqref{eqn:model}. Thus, we try to find $\dtree$, $\utree$, and $\ssent$ that are consistent with $\dsent$ (a partial observation of $\ssent$).

\begin{table}[]
\resizebox{\columnwidth}{!}{
\begin{tabular}{|cc|cc|cc|cc|cc|}
\hline 
 \multicolumn{2}{|c|}{\depf{parataxis}}   & \multicolumn{2}{c|}{\depf{appos}}    & \multicolumn{2}{c|}{\depf{list} }         &  \multicolumn{2}{c|}{\depf{advcl} }      & \multicolumn{2}{c|}{ \depf{ccomp} }           \\ 
   \multicolumn{2}{|c|}{\depf{2.38}}   & \multicolumn{2}{c|}{\depf{2.29}}     & \multicolumn{2}{c|}{\depf{1.33}}         & \multicolumn{2}{c|}{\depf{0.77}}           & \multicolumn{2}{c|}{\depf{0.53} }         \\ \hline
  \textbf{\litf{, ,}} &\textbf{26.8} & \textbf{\litf{, ,}} &\textbf{18.8}        & \litf{$\epsilon\ \epsilon$}& 60.0   &\litf{$\epsilon\ \epsilon$}& 73.8    & \litf{$\epsilon\ \epsilon$} &90.8     \\
  \litf{$\epsilon\ \epsilon$} &20.1  & \litf{:\ $ \epsilon$}&18.1  & \textbf{\litf{, ,}}& \textbf{22.3}   & \textbf{\litf{, ,}} &\textbf{21.2}   & \textbf{\litf{`` ''}}& \textbf{2.4  } \\
 \textbf{\litf{( )}}&\textbf{13.0}  & \litf{- $\epsilon$}&15.9       & \litf{,\ $\epsilon$}&5.3   & \litf{$\epsilon$ ,} &3.1     & \textbf{\litf{, ,}} &\textbf{2.4 }     \\
  \litf{-\ $\epsilon$} &9.7  & \litf{$\epsilon\ \epsilon$}& 14.4 & \textbf{\litf{< >} }&\textbf{3.0}   & \textbf{\litf{( )}} &\textbf{0.74 }    & \litf{:`` ''}& 0.9  \\
  \litf{:\ $\epsilon$} &8.1  & \textbf{\litf{( )}}& \textbf{13.1} & \textbf{\litf{( )}} &\textbf{3.0}   & \litf{$\epsilon$ -} &0.21     & \litf{`` ,''} &0.8   \\
\end{tabular}}
\caption{\label{tab:top} The top 5 relations that are most likely to generate symmetric punctemes, the entropy of their puncteme pair (row 2), and their top 5 puncteme pairs (rows 3--7) with their probabilities shown as percentages. The symmetric punctemes are in boldface.}
\end{table} 

The first step is to reconstruct $\dtree$ from $\dsent$.  This initial parsing step is intended to choose the $\dtree$ that maximizes $\psyn(\dtree \mid \dsent)$.\footnote{Ideally, rather than maximize, one would integrate over possible trees $\dtree$, in practice by sampling many values $\dtree_k$ from $\psyn(\cdot \mid \usent)$ and replacing $S(T)$ in \eqref{eqn:mbr} with $\bigcup_k S(T_k)$.} This step depends only on $\psyn$ and not on our punctuation model ($p_\attp$,
  $p_\ncp$).
  In practice, we choose $\dtree$ via a dependency parser that has been trained on an unpunctuated treebank with examples of the form $(\dsent,\dtree)$.\footnote{Specifically, the Yara parser \cite{rasooli-tetrault-2015}, a fast non-probabilistic transition-based parser that uses rich non-local features \cite{zhang-nivre:2011:ACL-HLT2011}.}  

\Cref{eqn:conditmodel} now defines a distribution over $(\utree,\spunc)$ 
given this $\dtree$.  To obtain a single prediction for $\spunc$, we adopt the minimum Bayes risk (MBR) approach of choosing surface punctuation $\hat{\spunc}$ that minimizes the expected loss with respect to the unknown truth $\spunc^*$.  Our loss function is the total edit distance over all slots (where edits operate on punctuation tokens).  Finding $\hat{\spunc}$ exactly would be intractable, so we use a sampling-based approximation and draw $m=1000$ samples from the posterior distribution over $(\utree,\spunc)$. We then define 

\noindent
\begin{align}\label{eqn:mbr}
    \hat{\spunc} = \argmin _{\spunc \in S(\dtree)} \sum_{\spunc^* \in S(\dtree)} \hat{p}(\spunc^* | \dtree) \cdot \text{loss}(\spunc,\spunc^*)
\end{align}
where $S(\dtree)$ is the set of unique $\spunc$ values in the sample and $\hat{p}$ is the empirical distribution given by the sample.  This can be evaluated in $O(m^2)$ time.

We evaluate on Arabic, English, Chinese, Hindi, and Spanish.  For each language, we train both the parser and the punctuation model on the training split of that UD treebank (\cref{sec:data}), and evaluate on held-out data.  We compare to the BiLSTM-CRF baseline in \cref{sec:baseline} \cite{bilstm-crf}.\footnote{We copied their architecture exactly but re-tuned the hyperparameters on our data.  We also tried tripling the amount of training data by adding unannotated sentences (provided along with the original annotated sentences by \newcite{11234/1-1989}), taking advantage of the fact that the BiLSTM-CRF does not require its training sentences to be annotated with trees.  However, this actually hurt performance slightly, perhaps because the additional sentences were out-of-domain.  We also tried the BiGRU-with-attention architecture of \citet{tilk2016bidirectional}, but it was also weaker than the BiLSTM-CRF (just as in \cref{tab:perp}).  We omit all these results from \cref{fig:restore} to reduce clutter.} 
We also compare to a ``trivial'' deterministic baseline, which merely places a period at the end of the sentence  (or a "\textbar" in the case of Hindi) and adds no other punctuation.  Because most slots do not in fact have punctuation, the trivial baseline already does very well; to improve on it, we must fix its errors without introducing new ones.
\definecolor{ar}{rgb}{0.62, 0.0, 0.77}
\definecolor{zh}{rgb}{0.82, 0.1, 0.26}
\definecolor{en}{rgb}{1.0, 0.49, 0.0}
\definecolor{hi}{rgb}{0.0, 0.5, 0.0}
\definecolor{es}{rgb}{0.0, 0.0, 1.0}
\definecolor{oracle}{rgb}{0.5, 0.5, 0.5}
\newcommand{\trsym}{\raisebox{-1pt}{{\tiny \DiamondSolid}}}
\newcommand{\bcsym}{\raisebox{-2pt}{\scriptsize \FiveStar}}
\newcommand{\prsym}{\raisebox{-1pt}{\tiny \CircleSolid}\raisebox{-1pt}{{}-{}-}}
\newcommand{\prnoisysym}{\raisebox{-1pt}{\tiny \Circle}\raisebox{-1pt}{{}-{}-}}
\newcommand{\ocsym}{\raisebox{-1pt}{{}-{}-}\raisebox{-1pt}{\tiny \CircleSolid}}
\newcommand{\restbl}{
\setlength\tabcolsep{2pt}
\scalebox{.7}{
\begin{tabular}{l|c|c|c||c|c|c}
    &\trsym &\bcsym & \ATT{}			 &\prsym                                        & \orcf{\ocsym}\\\hline
    \bf{\color{ar}Arabic}  &0.064&0.064& 0.063	         & \textbf{0.059}            & \orcf{0.053} \\
    \bf{\color{zh}Chinese} &0.110&\textbf{0.109} & \textbf{0.104}  & \textbf{0.102}  & \orcf{0.048} \\
    \bf{\color{en}English}&0.100 &0.108& 0.092		     & \textbf{0.090}            & \orcf{0.079} \\
    \bf{\color{hi}Hindi}&0.025&0.023& \textbf{0.019}  & \textbf{0.018}               & \orcf{0.013} \\
    \bf{\color{es}Spanish}&0.093&0.092& 0.085		     & \textbf{0.078}             & \orcf{0.068}	\\
\end{tabular}}
}

\begin{figure}[t]
\scalebox{.9}{
\hspace{-.6em}\begin{tikzpicture}
\tikzstyle{every node}=[font=\large]
    \node[draw=none] at (0,0) {\includegraphics[width=\columnwidth]{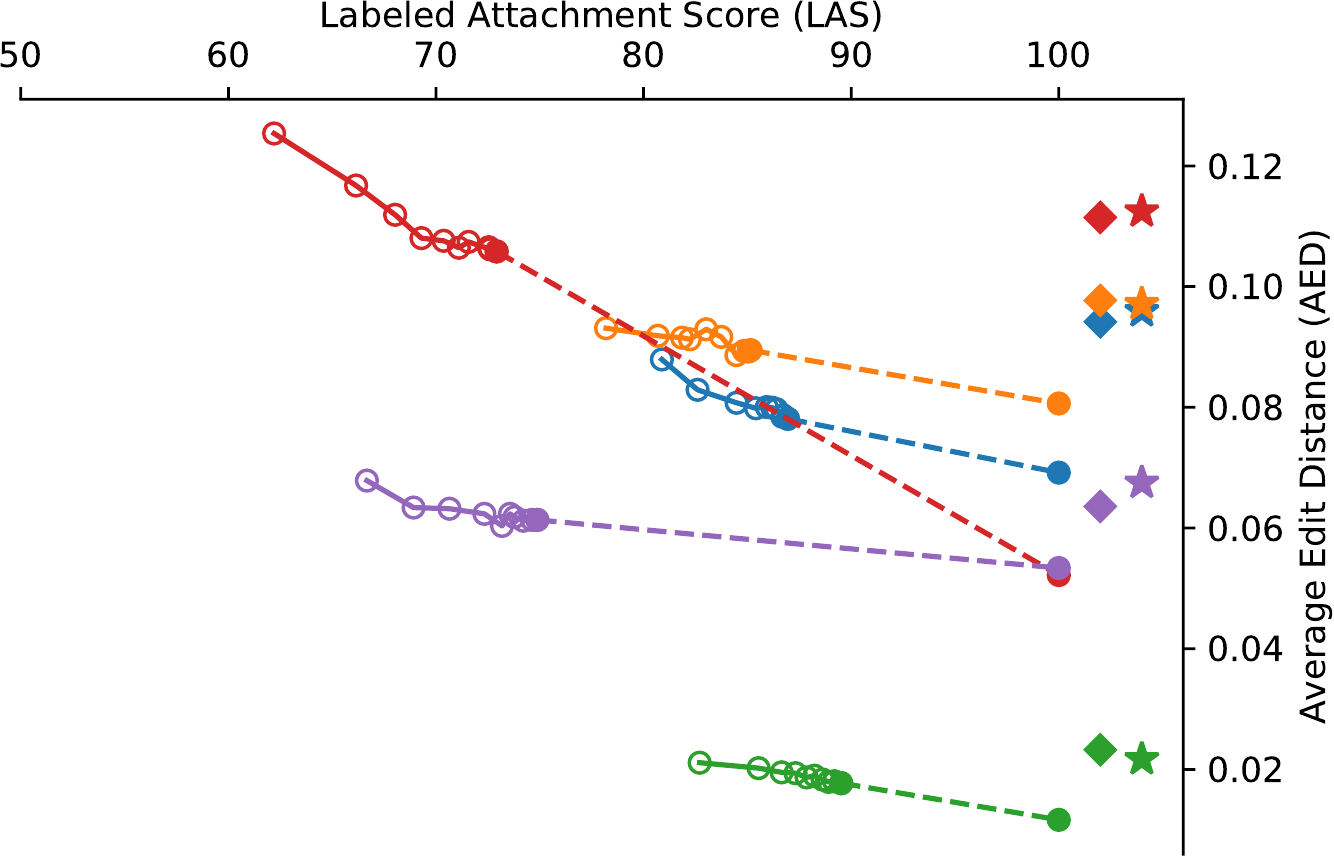}};
    \node[draw=none,inner sep=0pt,outer sep=0pt] at (-1.9,-1.6) {\restbl};
\end{tikzpicture}
}
\caption{\label{fig:restore}Edit distance per slot (which we call average edit distance, or AED) for each of the 5 corpora.  Lower is better.  The table gives the final AED on the {\bf test} data.  Its first 3 columns show the baseline methods just as in \cref{tab:perp}: the trivial deterministic method, the BiLSTM-CRF, and the \ATT{} ablation baseline that attaches the surface punctuation directly to the tree.  Column 4 is our method that incorporates a noisy channel, and column 5 (in gray) is our method using oracle (gold) trees.  We boldface the best non-oracle result as well as all that are not significantly worse (paired permutation test, $p < 0.05$).  The curves show how our method's AED (on {\bf dev} data) varies with the labeled attachment score (LAS) of the trees, where \orcf{\ocsym} at $x=100$ uses the oracle (gold) trees, \prsym{} at $x < 100$ uses trees from our parser trained on 100\% of the training data, and the \prnoisysym{} points at $x \ll 100$ use increasingly worse parsers.  The \trsym{} and \bcsym{} at the right of the graph show the AED of the trivial deterministic baseline and the BiLSTM-CRF baseline, which do not use trees.}
\end{figure}

Our final comparison on test data is shown in the table in \cref{fig:restore}.  On all 5 languages, our method beats (usually significantly) its 3 competitors: the trivial deterministic baseline, the BiLSTM-CRF, and the ablated version of our model (\ATT) that omits the noisy channel.

Of course, the success of our method depends on the quality of the parse trees $\dtree$ (which is particularly low for Chinese and Arabic).  The graph in \cref{fig:restore} explores this relationship, by evaluating (on dev data) with noisier trees obtained from parsers that were variously trained on only the first 10\%, 20\%, \ldots of the training data.  On all 5 languages, provided that the trees are at least 75\% correct, our punctuation model beats both the trivial baseline and the BiLSTM-CRF (which do not use trees).
It also beats the \ATT{} ablation baseline at all levels of tree accuracy (these curves are omitted from the graph to avoid clutter).  In all languages, better parses give better performance, and gold trees yield the best results.

\newcommand{\underlying}{\operatorname{underlying}}
\newcommand{\CESL}{\operatorname{CESL}}
\newcommand{\ESL}{\operatorname{ESL}}
\newcommand{\NCESL}{\operatorname{NC-ESL}}
\newcommand{\NCCESL}{\operatorname{NC-CESL}}
\newcommand{\edge}{\operatorname{edge}}
\subsection{Punctuation Correction}
\label{sec:corr}
Our next goal is to correct punctuation errors in a learner
corpus.
Each sentence is drawn from the Cambridge Learner Corpus treebanks, which provide original (\tbf{en\_esl}) and corrected (\tbf{en\_cesl}) sentences. 
All kinds of errors are corrected, such as syntax errors, but we use only the 30\% of sentences whose depunctuated trees $\dtree$ are isomorphic between \tbf{en\_esl} and \tbf{en\_cesl}.  These \tbf{en\_cesl} trees may correct word and/or punctuation errors in \tbf{en\_esl}, as we wish to do automatically.

We assume that an English learner can make mistakes in both the attachment and the noisy channel steps.  A common attachment mistake is the failure to
surround a non-restrictive relative clause with commas. In the noisy channel step, mistakes in quote transposition are common.\cutforspace{A further step would be to model typos by adding another noisy channel, which we didn't implement in this experiment.}

\newcommand{\sserr}{\ssent_e}
\newcommand{\sscor}{\ssent_c}
\newcommand{\xxcor}{\spunc_c}
\newcommand{\xxerr}{\spunc_e}  
\newcommand{\uucor}{\upunc_c}
\newcommand{\uuerr}{\upunc_e}
\newcommand{\ddcor}{\dsent_c}
\newcommand{\dderr}{\dsent_e}

\paragraph{Correction model.}
Based on the assumption about the two error sources, we develop a discriminative model for this task.  Let $\sserr$ denote the full input sentence, and let $\xxerr$ and $\xxcor$ denote the input (possibly errorful) and output (corrected) punctuation sequences.  
We model $p(\xxcor\mid \sserr) = \sum_{\dtree} \sum_{\utree_c} \psyn(\dtree\mid \sserr) \cdot p_\attp(\utree_c\mid \dtree,\xxerr) \cdot p_\ncp (\xxcor\mid \utree_c)$.  Here $\dtree$ is the depunctuated parse tree, $\utree_c$ is the corrected underlying tree, $\utree_e$ is the error underlying tree, and we assume
\mbox{$p_\attp(\utree_c\mid \dtree,\xxerr) = \sum_{\utree_e} p(\utree_e\mid \dtree,\xxerr) \cdot p_\attp(\utree_c \mid \utree_e)$}. 

In practice we use a 1-best pipeline rather than summing.  
Our first step is to reconstruct $\dtree$ from the error sentence $\sserr$. We choose $\dtree$ that maximizes $\psyn(\dtree\mid \sserr)$ from a dependency parser trained on \tbf{en\_esl} treebank examples ($\sserr$, $\dtree$). 
The second step is to reconstruct $\utree_e$ based on our punctuation model trained on \tbf{en\_esl}. We choose $\utree_e$ that maximizes $p(\utree_e\mid \dtree,\xxerr)$.
We then reconstruct $\utree_c$ by 

\noindent
\begin{align}
\p(\utree_c \mid \utree_e) = \textstyle{\prod_{\node_{e} \in \utree_e}}\;\p(\lpuncteme, \rpuncteme \mid \node_{e})
\end{align}
where $\node_{e}$ is the node in $\utree_{e}$, and $\p(\lpuncteme, \rpuncteme \mid \node_{e})$ is a similar log-linear model to \cref{eq:loglin} with additional features (\appref{sec:template_corr}\footnoteref{ft:supp})
which look at $\node_{e}$.

Finally, we reconstruct $\xxcor$ based on the noisy channel $p_\ncp (\xxcor\mid \utree_c)$ in \cref{sec:nc}. During training, $\ncp$ is regularized to be close to the noisy channel parameters in the punctuation model trained on \tbf{en\_cesl}.

We use the same MBR decoder as in \cref{sec:restore} to choose the best action. We evaluate using AED as in \cref{sec:restore}.  As a second metric, we use the script from the CoNLL 2014 Shared Task on Grammatical Error Correction \cite{W14-1701}: it computes the F$_{0.5}$-measure of the set of edits found by the system, relative to the true set of edits.

As shown in \cref{table:correct}, our method achieves better performance than the punctuation restoration baselines (which ignore input punctuation).  On the other hand, it is soundly beaten by a new BiLSTM-CRF that we trained specifically for the task of punctuation correction.  This is the same as the BiLSTM-CRF in the previous section, except that the BiLSTM now reads a \emph{punctuated} input sentence (with possibly erroneous punctuation)\cutforspace{ instead of an unpunctuated one}.  To be precise, at step $0 \leq i \leq n$, the BiLSTM reads a concatenation of the embedding of word $i$ (or $\bos$ if $i=0$) with an embedding of the punctuation token sequence $\sslot_i$.
The BiLSTM-CRF wins because it is a discriminative model tailored for this task: the BiLSTM can extract arbitrary contextual features of slot $i$ that are correlated with whether $\sslot_i$ is correct in context.

\begin{table}
\centering
\begin{tabular}{c|c|c|c|c|c|c}
   &\trsym          & \bcsym   &\prsym      & parsed      &   \orcf{gold}   & \bcsym-corr   \\ \hline
AED&    0.052   & 0.051   & 0.047    & 0.034   &   \orcf{0.033} &  \textbf{0.005} \\
F$_{0.5}$ & 0.779 & 0.787 &0.827 & 0.876 & 0.881 & \textbf{0.984}
\end{tabular}
\caption{\label{table:correct} AED and F$_{0.5}$ results on the test split of English-ESL data.  Lower AED is better; higher F$_{0.5}$ is better.  The first three columns (markers correspond to \cref{fig:restore}) are the punctuation restoration baselines, which ignore the input punctuation. The fourth and fifth columns are our correction models, which use parsed and \orcf{gold} trees. The final column is the BiLSTM-CRF model tailored for the punctuation correction task.}
\end{table}

\subsection{Sentential Rephrasing}
\label{sec:gd}

We suspect that syntactic transformations on a sentence should often preserve the underlying punctuation attached to its tree.  The surface punctuation can then be regenerated from the transformed tree.  Such transformations include edits that are suggested by a writing assistance tool \cite{heidorn2000intelligent}, or subtree deletions in compressive summarization \cite{knight2002summarization}.

For our experiment, we evaluate an interesting case of syntactic transformation. \citet{wang-eisner-2016} consider a systematic rephrasing procedure by rearranging the order of dependent subtrees within a UD treebank, in order to synthesize new languages with different word order that can then be used to help train multi-lingual systems (i.e., data augmentation with synthetic data).%

As \citeauthor{wang-eisner-2016} acknowledge (\citeyear[footnote~9]{wang-eisner-2016}), their permutations treat surface punctuation tokens like ordinary words, which can result in synthetic sentences whose punctuation is quite unlike that of real languages. 

\newcommand{\udbfr}{
\hspace{-5pt}\begin{dependency}[baseline=-.3em,label style={draw=none,inner sep=0ex,outer sep=0ex},edge slant=5pt,edge vertical padding=-5pt]
\begin{deptext}[column sep=.05em, row sep=-.5em]
\posf{\tiny SCONJ}\&\posf{\tiny ADJ}\&\posf{\tiny PUNCT}\&\posf{\tiny DET}\&\posf{\tiny NOUN}\&\posf{\tiny VERB}\&\posf{\tiny PUNCT}\\
\litf{If}\&\litf{true}\&\litf{,}\&\litf{the}\&\litf{caper}\&\litf{failed}\&\litf{.}\\
\end{deptext}
\depedge[edge unit distance=1ex]{2}{1}{\depf{mark}}
\depedge[edge unit distance=.3ex]{5}{4}{\depf{det}}
\depedge[edge unit distance=.6ex]{6}{3}{\depf{punct}}
\depedge[edge unit distance=.7ex]{6}{2}{\depf{advcl}}
\depedge[edge unit distance=.3ex]{6}{5}{\depf{nsubj}}
\depedge[edge unit distance=.5ex]{6}{7}{\depf{punct}}
\deproot[edge unit distance=.85ex]{6}{\depf{root}}
\end{dependency}\hspace{-5pt}
}
\newcommand{\udaft}{
\hspace{-5pt}\begin{dependency}[baseline=-.3em,label style={draw=none,inner sep=0ex,outer sep=0ex},edge slant=5pt,edge vertical padding=-5pt]
\begin{deptext}[column sep=.05em, row sep=-.5em]
\posf{\tiny DET}\&\posf{\tiny NOUN}\&\posf{\tiny VERB}\&\posf{\tiny PUNCT}\&\posf{\tiny SCONJ}\&\posf{\tiny ADJ}\&\posf{\tiny PUNCT}\\
\litf{the}\&\litf{caper}\&\litf{failed}\&\litf{.}\&\litf{If}\&\litf{true}\&\litf{,}\\
\end{deptext}
\depedge[edge unit distance=.3ex]{6}{5}{\depf{mark}}
\depedge[edge unit distance=.6ex]{2}{1}{\depf{det}}
\deproot[edge unit distance=.7ex]{3}{\depf{root}}
\depedge[edge unit distance=.6ex]{3}{2}{\depf{nsubj}}
\depedge[edge unit distance=.3ex]{3}{4}{\depf{punct}}
\depedge[edge unit distance=.4ex]{3}{6}{\depf{advcl}}
\depedge[edge unit distance=.6ex]{3}{7}{\depf{punct}}
\end{dependency}\hspace{-5pt}
}

In our experiment, we use \citeposs{wang-eisner-2016} ``self-permutation'' setting, where the dependents of each noun and verb are stochastically reordered, but according to a dependent ordering model that has been trained on the same language.  For example, rephrasing a English sentence \udbfr under an English ordering model may yield \\ \udaft which is still grammatical except that \litf{,} and \litf{.} are wrongly swapped (after all, they have the same POS tag and relation type).
Worse, permutation may yield bizarre punctuation such as \litf{, ,} at the start of a sentence.

\newcommand{\underlybfr}{
\hspace{-5pt}\begin{dependency}[baseline=-.3em,label style={draw=none,inner sep=0ex,outer sep=.5ex},edge vertical padding=-3pt]
\begin{deptext}[column sep=.2em, row sep=0em]
    \texttt{\undb{\bossym,}}\&\hspace{-6pt}\texttt{If}\&\texttt{true\undb{,}}\&\texttt{the}\&\texttt{caper}\&\texttt{failed}\&\hspace{-6pt}\texttt{\undb{.}}\\
\end{deptext}
\depedge[edge unit distance=.6ex]{5}{4}{\texttt{det}}
\depedge[edge unit distance=.6ex]{6}{5}{\texttt{nsubj}}
\deproot[edge unit distance=.8ex,label style={outer sep=.5ex}]{6}{\hspace{5pt}\texttt{root}\undb{\texttt{.}}}
\depedge[edge unit distance=1ex]{3}{2}{\texttt{mark}}
\depedge[edge unit distance=.5ex,label style={above}]{6}{3}{\undb{\texttt{,}}\texttt{advcl}\undb{\texttt{,}}}
\end{dependency}\hspace{-5pt}
}
\newcommand{\underlyaft}{
\hspace{-5pt}\begin{dependency}[baseline=-.3em,label style={draw=none,inner sep=0ex,outer sep=.5ex},edge vertical padding=-3pt]
\begin{deptext}[column sep=.2em, row sep=0em]
\texttt{\undb{\bossym}the}\&\texttt{caper}\&\texttt{failed}\&\texttt{\undb{,}If}\&\texttt{true}\&\hspace{-6pt}\texttt{\undb{,.}}\\
\texttt{\undr{\bossym}the}\&\texttt{caper}\&\texttt{failed}\&\texttt{\undr{,}If}\&\texttt{true}\&\hspace{-6pt}\texttt{\undr{.}}\\
\end{deptext}
\depedge[edge unit distance=.8ex]{2}{1}{\texttt{det}}
\depedge[edge unit distance=.8ex]{3}{2}{\texttt{nsubj}}
\deproot[edge unit distance=.8ex,label style={outer sep=.5ex}]{3}{\hspace{5pt}\texttt{root}\undb{\texttt{.}}}
\depedge[edge unit distance=.6ex]{5}{4}{\texttt{mark}}
\depedge[edge unit distance=.7ex,label style={above}]{3}{5}{\undb{\texttt{,}}\texttt{advcl}\undb{\texttt{,}}}
\end{dependency}\hspace{-5pt}
}

Our punctuation model gives a straightforward remedy---instead of permuting the tree directly, we first discover its most likely underlying tree \underlybfr \\ by the maximizing variant of \cref{alg:inside}  (\cref{sec:algo}). Then, we permute the underlying tree and sample the surface punctuation from the distribution modeled by the trained PFST,
yielding \\
\underlyaft 
We leave the handling of capitalization to future work.

We test the naturalness of the permuted sentences by asking how well a word trigram language model trained on them could predict the original sentences.\footnote{So the two approaches to permutation yield different training data, but are compared fairly on the same test data.
}
As shown in \cref{tb:gd}, our permutation approach reduces the perplexity over the baseline on 4 of the 5 languages, often dramatically.

\begin{table}
\setlength\tabcolsep{2pt}
\centering
\begin{tabular}{l|ccc|ccc}
           &\multicolumn{3}{c|}{Punctuation}&\multicolumn{3}{c}{All}       \\\hline
 & Base     &Half& Full           & Base      &Half& Full          \\\hline
Arabic     & {\bf 156.0}  &231.3& 186.1         & {\bf 540.8}&590.3 & 553.4        \\
Chinese    & 165.2        &110.0& {\bf 61.4 }   & 205.0      &174.4& {\bf 78.7 }  \\
English    & 98.4         &74.5& {\bf 51.0 }   & 140.9       &131.4 & {\bf 75.4 }  \\
Hindi      & 10.8         &11.0& {\bf 9.7  }   & 118.4       &118.8& {\bf 91.8 }  \\
Spanish    & 266.2        &259.2& {\bf 194.5}   & 346.3      &343.4 & {\bf 239.3}  \\
\end{tabular}
\caption{\label{tb:gd}Perplexity (evaluated on the train split to avoid evaluating generalization) of a trigram language model trained (with add-$0.001$ smoothing) on different versions of \emph{rephrased training sentences}. ``Punctuation'' only evaluates
perplexity on the trigrams that have punctuation. ``All'' evaluates on all the trigrams. ``Base'' permutes all surface dependents including punctuation \citep{wang-eisner-2016}.  ``Full'' is our full approach: recover underlying punctuation, permute remaining dependents, regenerate surface punctuation.  ``Half'' is like ``Full'' but it permutes the non-punctuation tokens identically to ``Base.''  The permutation model is trained on surface trees or recovered underlying trees $\utree$, respectively.  In each 3-way comparison, we boldface the best result (always significant under a paired permutation test over per-sentence log-probabilities, $p<0.05$).}
\end{table}

\section{Related Work}

Punctuation can aid syntactic analysis, since it signals phrase boundaries and sentence structure.
\citet{briscoe1994parsing} and \citet{white-rajkumar:2008:GEAF}
parse punctuated sentences using hand-crafted constraint-based grammars that implement Nunberg's approach in a declarative way.  These grammars treat \emph{surface} punctuation symbols as ordinary words, but annotate the nonterminal categories so as to effectively keep track of the \emph{underlying} punctuation.  This is tantamount to crafting a grammar for underlyingly punctuated sentences and composing it with a finite-state noisy channel.  

The parser of \citet{ji-zhang-zhu-2014} takes a different approach and treats punctuation marks as features of their neighboring words.  \citet{parsing-gen-punc} use a generative model for punctuated sentences, leting them restore punctuation marks during transition-based parsing of unpunctuated sentences. \citet{I05-2002} use punctuation marks to segment a sentence: this "divide and rule" strategy reduces ambiguity in parsing of long Chinese sentences.  Punctuation can similarly be used to constrain syntactic structure during grammar induction \cite{Spitkovsky:2011:PMP:2018936.2018939}.\cutforspace{  Punctuation has been used to remove redundant treebank rules, yielding cleaner training data \cite{tse-curran-2008}.}

Punctuation restoration (\cref{sec:restore}) is useful for transcribing text from unpunctuated speech. The task is usually treated by tagging each slot with zero or more punctuation tokens, using a traditional sequence labeling method: conditional random fields \cite{restore-seq-labeling,lu2010better}, recurrent neural networks \cite{tilk2016bidirectional}, or transition-based systems \cite{mult-lingual-gen}. 

\section{Conclusion and Future Work}
\label{sec:future}

We have provided a new computational approach to modeling punctuation.  In our model, syntactic constituents stochastically generate latent underlying left and right punctemes.  Surface punctuation marks are not directly attached to the syntax tree, but are generated from sequences of adjacent punctemes by a (stochastic) finite-state string rewriting process  .
Our model is inspired by \citeposs{nunberg1990linguistics} formal grammar for English punctuation, but is probabilistic and trainable.  We give exact algorithms for training and inference.

We trained Nunberg-like models for 5 languages and L2 English. We compared the English model to Nunberg's, and showed how the trained models can be used across languages for punctuation restoration, correction, and adjustment.

In the future, we would like to study the usefulness of the recovered underlying trees on tasks such as syntactically sensitive sentiment analysis \cite{P15-1150}, machine translation \cite{W06-1628}, relation extraction \cite{P04-1054}, and coreference resolution \cite{C10-1068}. We would also like to investigate how underlying punctuation could aid parsing. For discriminative parsing, features for scoring the tree could refer to the underlying punctuation, not just the surface punctuation. For generative parsing (\cref{sec:inference}), we could follow the scheme in \cref{eqn:model}.
For example, the $\psyn$ factor in \cref{eqn:model} might be a standard recurrent neural network grammar (RNNG) \cite{N16-1024}; when a subtree of $\dtree$ is completed by the \textsc{Reduce} operation of $\psyn$, the punctuation-augmented RNNG~\eqref{eqn:model} would stochastically attach subtree-external left and right punctemes with $p_\attp$ and transduce the subtree-internal slots with $p_\ncp$.

In the future, we are also interested in enriching the $\utree$ representation and making it more different from $\dtree$, to underlyingly account for other phenomena in $\dtree$ such as capitalization, spacing, morphology, and non-projectivity (via reordering).

\section*{Acknowledgments} 
This material is based upon work supported by the National Science Foundation under Grant Nos.\@ 1423276 and 1718846, including a REU supplement to the first author.  We are grateful to the state of Maryland for the Maryland Advanced Research Computing Center, a crucial resource. We thank Xiaochen Li for early discussion, Argo lab members for further discussion, and the three reviewers for quality comments.\par

\bibliography{latstruct}
\bibliographystyle{acl_natbib}

\clearpage\appendix\appendixpage\pagestyle{empty}
\newcommand{\zhpair}{\zh{《》}\zh{〈〉}\zh{【】}\zh{『』}\zh{「」}}
\section{Feature Templates for \ATT{}}
\label{sec:template}

Below, we provide \defn{feature templates} for the features used by the \textsc{Attach} model in \cref{sec:gen}.  To illustrate, \cref{tb:feature} lists all the non-backoff features that fire on a particular node in \cref{fig:gen_story} of the main paper.  Specifically, \cref{tb:feature} lists the nonzero features in the feature vector $\vf(\lpuncteme,\rpuncteme,\node)$ where $\node$ is the tree node that dominates the subject \texttt{Dale} and $(\lpuncteme,\rpuncteme)=(\texttt{``},\texttt{''})$ says to surround that subject with quotation marks.

In general, the feature vector $\vf(\lpuncteme,\rpuncteme,\node)$ assigns nonzero values (1 values unless otherwise stated) to the features that are named by the following tuples.  (We use dots here to separate the elements of a tuple.)

\begin{itemize}
    \item $\texttt{N}.\lpuncteme.\rpuncteme.\pos.\bar{\dep}$, $\texttt{N}.\lpuncteme.\rpuncteme.\pos.\dep$, $\texttt{N}.\lpuncteme.\rpuncteme.\pos$, $\texttt{N}.\lpuncteme.\rpuncteme.\bar{\dep}$, and $\texttt{N}.\lpuncteme.\rpuncteme.\dep$, where $\pos$ is the POS-tag of the word at $\node$, $\dep$ is the dependency relation that labels the edge of $\dtree$ that points to $\node$, and $\bar{\dep} = \uphalfleftarrow{\dep}$ or $\uphalfrightarrow{\dep}$ according to the direction of that edge.  The first feature name is most specific, while the remaining 4 features are {\bf backoff features}.

      For example, such features can be used to say that an appositive ($d=\depf{appos}$) headed by a noun ($g=\posf{NOUN}$) likes to be surrounded by commas ($\lpuncteme=\rpuncteme=\litf{,}$).

      To make training faster and perhaps avoid local optima, we initialize the weight of feature $\texttt{N}.l.r.d$ to its log-count in training data.

    \item $\texttt{W}.h.\lpuncteme.\rpuncteme.\pos.\bar{\dep}$, $\texttt{W}.h.\lpuncteme.\rpuncteme.\pos.\dep$, $\texttt{W}.h.\lpuncteme.\rpuncteme.\pos$, $\texttt{W}.h.\lpuncteme.\rpuncteme.\bar{\dep}$, and $\texttt{W}.h.\lpuncteme.\rpuncteme.\dep$, where $h$ measures the length of the constituent headed by $\node$: $h=1$ for a short constituent (1--2 words), $h=2$ for a medium constituent (3--5 words), and $h=3$ for a long constituent ($\geq 6$ words).

      For example, a positive weight on \texttt{W}.\litf{3}.\litf{,}.\litf{,}.\depf{advcl} says that long subordinate clauses ($h=3$, $d=\depf{advcl}$) are likely to be surrounded by commas.

\item $\texttt{A}.\lpuncteme.\rpuncteme.\pos.\bar{\dep}.\dep'$, $\texttt{A}.\lpuncteme.\rpuncteme.\pos.\dep.\dep'$, $\texttt{A}.\lpuncteme.\rpuncteme.\pos.\dep'$, $\texttt{A}.\lpuncteme.\rpuncteme.\bar{\dep}.\dep'$, and $\texttt{A}.\lpuncteme.\rpuncteme.\dep.\dep'$, for each dependency relation $\dep'$ that occurs along the path from the root of $\dtree$ to the parent of $\node$.  (Here $\lpuncteme,\rpuncteme$, and $\pos$ are properties of $\node$ as before, whereas $\dep'$ refers to an ancestor of $\node$.)  The value of this feature is the number of times that $\dep'$ appears along the path.  Notice that if $\dep=\depf{root}$, the path is empty, so none of the $\texttt{A}$ features fire.

  For example, such features might cause a subordinate clause to be punctuated differently depending on whether it is attached to the main verb or a more deeply nested verb.

\item $\texttt{C}.\lpuncteme.\rpuncteme.\pos.\bar{\dep}.\dep'$, $\texttt{C}.\lpuncteme.\rpuncteme.\pos.\dep.\dep'$, $\texttt{C}.\lpuncteme.\rpuncteme.\pos.\dep'$, $\texttt{C}.\lpuncteme.\rpuncteme.\bar{\dep}.\dep'$, and $\texttt{C}.\lpuncteme.\rpuncteme.\dep.\dep'$, for each 
  dependency relation $\dep'$ that appears on an edge from $\node$ to a child of $\node$.  The value of this feature is the number of such edges.  Notice that if $\node$ is a leaf, it has no children, so none of the \texttt{C} features fire.

  For example, such features could be used to say that a relative clause that contains a subject ($\dep'=\depf{subj}$), such as an object-relative clause, likes to be surrounded by commas.

\item $\texttt{L}.\lpuncteme.\pos_{-1}.\pos_{+1}$ and $\texttt{R}.\rpuncteme.\pos_{-1}.\pos_{+1}$, where $\pos_{-1}$ and $\pos_{+1}$ are the POS-tags surrounding the slot where $\lpuncteme$ or $\rpuncteme$ (respectively) is generated.  We use $\pos_{-1}=\bos$ or $\pos_{+1}=\eos$ if the slot is at the beginning or the end of the sentence (respectively).

\item $\texttt{S}.\pos.\bar{\dep}$, $\texttt{S}.\pos.\dep$, $\texttt{S}.\pos$, $\texttt{S}.\bar{\dep}$ and $\texttt{S}.\dep$, provided that $\lpuncteme$ and $\rpuncteme$ are \textbf{symmetric punctemes}.  Symmetry is determined by simultaneously scanning $\lpuncteme$ from left to right and $\rpuncteme$ from right to left, and checking whether the punctuation marks at each position form one of the following pairs:\footnote{A more complete list could be compiled from Unicode's opening/closing punctuation pairs, but this list is sufficient for the experiments in this paper.} \texttt{\{\} \hspace{-3pt}[] \hspace{-3pt}() \hspace{-3pt}``'' \hspace{-3pt}<> \hspace{-3pt}?`? \hspace{-3pt}!`!  \zhpair \hspace{-8pt},, \hspace{-3pt}-{}- }.  If $l$ and $r$ are both empty strings, they are not considered symmetric.

\item $c.\lpuncteme.\rpuncteme.\pos.\bar{\dep}$, $c.\lpuncteme.\rpuncteme.\pos.\dep$, $c.\lpuncteme.\rpuncteme.\pos$, $c.\lpuncteme.\rpuncteme.\bar{\dep}$ and $c.\lpuncteme.\rpuncteme.\dep$, for each punctuation token $c$ that appears at least once as \emph{surface} punctuation within the constituent dominated by $\node$.  (That is, if $\node$'s constituent stretches from slot $i$ to slot $k$, its \defn{internal slots} are $j = i+1,\ldots,k-1$, and $c$ must appear in $\sslot_j$ for some such $j$.)

  These features make it possible to implement punctuation marks of different precedence.  For example, a conjunct is ordinarily delimited by commas (\cref{sec:underly}), but a conjunct that already contains internal commas ($c=\litf{,}$) may be delimited by semicolons instead, as shown below.\footnote{Unfortunately, this feature does not explain why all \emph{other} conjuncts in the same conjunction (including the final conjunct) also switch to semicolons.}  Similarly, an appositive that already contains internal commas may be delimited by dashes instead of commas.
  \begin{quote}
    \scriptsize
    \hspace{-2em}\litf{There are two ways to read newspapers: }
    \litf{\ \ in print, which is costly; }
    \litf{\ \ \ \ or in digital, which is free.}
  \end{quote}

\end{itemize}

Some of these features are not edge-local.  They look at entire paths or constituents, or the surface punctuation of a constituent.  However, they do admit tractable exact algorithms, similarly to a neural HMM \cite{W16-5907}.  How?

During training, \cref{alg:line:p} of \cref{alg:inside} is able to compute each feature vector $\vf(\lpuncteme,\rpuncteme,\node)$ given the observed input tree $\dtree$ and surface punctuation $\spunc$.  

\cref{sec:corr} and \cref{sec:gd} both need to find the 1-best underlying tree $\utree$ that corresponds to the given $\dtree$ and $\spunc$ of a treebank sentence, so that it can correct or permute that sentence.  As discussed at the end of \cref{sec:algo}, this makes use of the same feature vectors $\vf(\lpuncteme,\rpuncteme,\node)$, and merely replaces the inside algorithm with a Viterbi decoding algorithm.

The situation is slightly more difficult at test time, when $\dtree$ is still observed, but the surface punctuation is not observed and must be sampled (\cref{sec:restore}).  However, we can still do exact joint sampling of $\utree$ and $\spunc$ by traversing $\dtree$ \emph{bottom-up}.  That is, after we have processed the child nodes of $\node$, we can process $\node$ by sampling $\sslot_j$ at the internal slots between its children (using \NC{}) and \emph{then} sampling $(\ell,r)$ at its external slots (using \ATT{}, which may depend on the $\sslot_j$ values via the $c$ features).  

\setlength\tabcolsep{1.5pt}
\begin{table}[t]
  \centering
\begin{tabular}{|c|c|c|}
    \hline
Feature Type&Name & Value\\
\hline\hline
&&\\[-2.5ex]
$\texttt{N}.\lpuncteme.\rpuncteme.\pos.\bar{\dep}$&\texttt{N}.\litf{``}.\litf{''}.\posf{NOUN}.$\uphalfleftarrow{\depf{nsubj}}$ & 1\\\hline
&&\\[-2.5ex]
$\texttt{W}.h.\lpuncteme.\rpuncteme.\pos.\bar{\dep}$&\texttt{W}.\texttt{1}.\litf{``}.\litf{''}.\posf{NOUN}.$\uphalfleftarrow{\depf{nsubj}}$ & 1\\\hline
&&\\[-2.5ex]
$\texttt{S}.\pos.\bar{\dep}$&\texttt{S}.\posf{NOUN}.$\uphalfleftarrow{\depf{nsubj}}$ & 1\\\hline
&&\\[-2.5ex]
$\texttt{A}.\lpuncteme.\rpuncteme.\pos.\bar{\dep}.\dep'$&\texttt{A}.\litf{``}.\litf{''}.\posf{NOUN}.$\uphalfleftarrow{\depf{nsubj}}$.\depf{root}&1\\\hline
$\texttt{L}.\lpuncteme.\pos_{-1}.\pos_{+1}$&\texttt{L}.\litf{``}.\posf{BOS}.\posf{NOUN}&1\\
$\texttt{R}.\rpuncteme.\pos_{-1}.\pos_{+1}$&\texttt{R}.\litf{''}.\posf{NOUN}.\posf{VERB}&1\\\hline
\end{tabular}
\caption{\label{tb:feature} A subset of the features that fire on the node with \depf{nsubj} in \cref{fig:gen_story}.}
\end{table}

\section{Posterior Regularization}
\label{sec:pr}

\Cref{eqn:obj} includes the expectation of $c(\utree)$, which counts the nodes in $\utree$ whose $l$ and $r$ punctemes contain any unmatched punctuation tokens.

We define a criterion to decide whether $l$ and $r$ are unmatched, based on this list of matched symmetric tokens: \texttt{\{\} \hspace{-3pt}[] \hspace{-3pt}() \hspace{-3pt}``'' \hspace{-3pt}<> \hspace{-3pt}?`? \hspace{-3pt}!`! \zhpair}.  This is the same list used by the \texttt{S} feature in \cref{sec:template}, except that it omits the pairs where the two tokens are equal (namely \texttt{-{}-} and \texttt{,,}).

First, we modify $l$ and $r$ to filter out tokens that do not appear in the list above.  We then check whether the modified $l$ and $r$ are symmetric punctemes in the sense of the \texttt{S} feature (\cref{sec:template}).  If not, we count the node as having unmatched punctuation.

\section{Correction Feature Templates}
\label{sec:template_corr}

For the correction model (\cref{sec:corr}), recall that we first find the 1-best underlyingly punctuated tree $\utree_e$ that explains a tree $\dtree$ along with its possibly erroneous or non-standard surface punctuation $\spunc_e$.  

We then use \ATT{} to generate corrected punctuation to attach to  $\dtree$.  At this step, it may be beneficial to condition on knowledge of the reconstructed underlying punctuation that we reconstructed in $\utree_e$.  Thus, we add the following 2 feature templates, which are extended versions of the \texttt{N} and \texttt{W} features in \cref{sec:template}.
In these templates for evaluating $\vf(\lpuncteme,\rpuncteme,\node)$ in a proposed $\utree$, $\lpuncteme'$ and $\rpuncteme'$ denote the left and right underlying punctemes attached to the corresponding node $\node_e$ in $\utree_e$.

\begin{itemize}
\item  $\texttt{N}.\lpuncteme.\rpuncteme.\pos.\bar{\dep}.\lpuncteme'.\rpuncteme'$, $\texttt{N}.\lpuncteme.\rpuncteme.\bar{\dep}.\lpuncteme'.\rpuncteme'$, $\texttt{N}.\lpuncteme.\rpuncteme.\pos.\lpuncteme'.\rpuncteme'$, $\texttt{N}.\lpuncteme.\rpuncteme.\lpuncteme'.\rpuncteme'$
\item $\texttt{W}.h.\lpuncteme.\rpuncteme.\pos.\bar{\dep}.\lpuncteme'.\rpuncteme'$,$\texttt{W}.h.\lpuncteme.\rpuncteme.\bar{\dep}.\lpuncteme'.\rpuncteme'$, $\texttt{W}.h.\lpuncteme.\rpuncteme.\pos.\lpuncteme'.\rpuncteme'$,
$\texttt{W}.h.\lpuncteme.\rpuncteme.\lpuncteme'.\rpuncteme'$
\end{itemize}

\clearpage
\section{PFST implementation}\label{sec:pfst}

\paragraph{Construct the PFST}

\newcounter{sarrow}
\newcommand\xrsquigarrow[1]{%
\stepcounter{sarrow}%
\begin{tikzpicture}[decoration=snake]
\node (\thesarrow) {\tiny$#1$};
\draw[->,decorate] (\thesarrow.south west) -- (\thesarrow.south east);
\end{tikzpicture}%
}

\newcommand{\fstclr}{yellow!50}
\newcommand{\tcl}[1]{\textcircled{#1}}
\newcommand{\tfs}{\tcl{\raisebox{.12pt}{\scalebox{.8}{\tcl{\raisebox{-.8pt}{\fstfsym}}}}}}

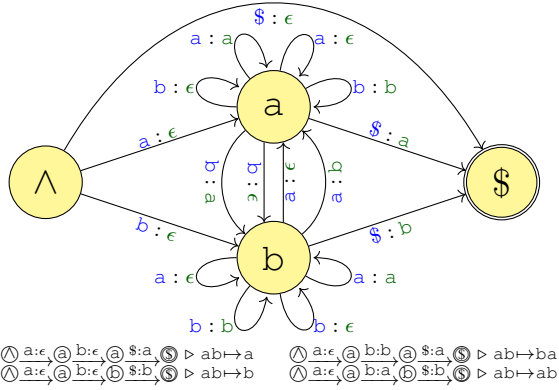
\begin{figure}[t]
\scalebox{1}{
\begin{tikzpicture}
\tikzstyle{every node}=[font=\scriptsize]
    [
      initial/.style={line width=1pt},
      accepting by double/.append style={line width=1pt},
      semithick,
    ]
\tikzstyle{vtx}=[fill=\fstclr]
    \draw node[vtx] (i) [state] at (0, 0) {\Large $\fstisym$};
    \draw node[vtx] (a) [state] at (3, 1) {\Large $\litf{a}$};
    \draw node[vtx] (b) [state] at (3, -1) {\Large $\litf{b}$};
    \draw node[vtx] (f) [state, accepting] at (6, 0){\Large $\fstfsym$};
    \path[->] (i) edge node[above=-3pt,sloped] {${\color{undb}\litf{a}}:{\color{undr}\epsilon}$} (a);
    \path[->] (i) edge node[below=-3pt,sloped] {${\color{undb}\litf{b}}:{\color{undr}\epsilon}$} (b);
    \path[->] (a) edge [in=150,out=180,loop] node[left=-3pt] {${\color{undb}\litf{b}}:{\color{undr}\epsilon}$} (a);
    \path[->] (a) edge [in=100,out=130,loop] node[left=-3pt,pos=.4] {${\color{undb}\litf{a}}:{\color{undr}\litf{a}       }$} (a);
    \path[->] (a) edge [in=50,out=80,loop]   node[right=-3pt,pos=.6] {${\color{undb}\litf{a}}:{\color{undr}\epsilon}$} (a);
    \path[->] (a) edge [in=0,out=30,loop]    node[right=-3pt] {${\color{undb}\litf{b}}:{\color{undr}\litf{b}       }$} (a);
    \path[->] (b) edge [in=210,out=180,loop] node[left=-3pt] {${\color{undb}\litf{a}}:{\color{undr}\epsilon}$} (b);
    \path[->] (b) edge [in=260,out=230,loop] node[left=-3pt,pos=.4] {${\color{undb}\litf{b}}:{\color{undr}\litf{b}       }$} (b);
    \path[->] (b) edge [in=310,out=280,loop] node[right=-3pt,pos=.6] {${\color{undb}\litf{b}}:{\color{undr}\epsilon}$} (b);
    \path[->] (b) edge [in=360,out=330,loop] node[right=-3pt] {${\color{undb}\litf{a}}:{\color{undr}\litf{a}       }$} (b);
    \path[->] (a.255) edge node[below=-3pt,sloped] {${\color{undb}\litf{b}}:{\color{undr}\epsilon}$} (b.105);
    \path[->] (a) edge [in=140,out=220] node[below=-3pt,sloped] {${\color{undb}\litf{b}}:{\color{undr}\litf{a}}$} (b);
    \path[->] (b.75) edge node[below=-3pt,sloped] {${\color{undb}\litf{a}}:{\color{undr}\epsilon}$} (a.-75);
    \path[->] (b) edge [in=320,out=40] node[below=-3pt,sloped] {${\color{undb}\litf{a}}:{\color{undr}\litf{b}}$} (a);
    \path[->] (a) edge  node[above=-3pt,sloped] {${\color{undb}\eoi}:{\color{undr}\litf{a}}$} (f);
    \path[->] (b) edge  node[below=-3pt,sloped] {${\color{undb}\eoi}:{\color{undr}\litf{b}}$} (f);
    \path[->] (i) edge [in=120,out=60,loop,distance=3cm] node[below=-2pt] {${\color{undb}\eoi}:{\color{undr}\epsilon}$} (f);
    \node[draw=none,text width=0cm] at (-.6,-2.27) {\scalebox{.85}{$\tcl{\fstisym}\ra{\litf{a}:\epsilon}\tcl{\litf{a}}\ra{\litf{b}:\epsilon}\tcl{\litf{a}}\ra{\eoi:\litf{a}}\tfs$ \comment{$\litf{ab}\myrarw\litf{a}$}}};
    \node[draw=none,text width=0cm] at (-.6,-2.53) {\scalebox{.85}{$\tcl{\fstisym}\ra{\litf{a}:\epsilon}\tcl{\litf{a}}\ra{\litf{b}:\epsilon}\tcl{\litf{b}}\ra{\eoi:\litf{b}}\tfs$ \comment{$\litf{ab}\myrarw\litf{b}$}}};
    \node[draw=none,text width=0cm] at (3.2,-2.27) {\scalebox{.85}{$\tcl{\fstisym}\ra{\litf{a}:\epsilon}\tcl{\litf{a}}\ra{\litf{b}:\litf{b}}\tcl{\litf{a}}\ra{\eoi:\litf{a}}\tfs$ \comment{$\litf{ab}\myrarw\litf{ba}$}}};
    \node[draw=none,text width=0cm] at (3.2,-2.53) {\scalebox{.85}{$\tcl{\fstisym}\ra{\litf{a}:\epsilon}\tcl{\litf{a}}\ra{\litf{b}:\litf{a}}\tcl{\litf{b}}\ra{\eoi:\litf{b}}\tfs$ \comment{$\litf{ab}\myrarw\litf{ab}$}}};
  \end{tikzpicture}
  }
  \caption{\label{fig:pfst} An example of our PFST on vocabulary $\pv = \{\litf{a}, \litf{b}\}$. The input (underlying punctuation tokens) is colored in {\color{undb}blue} and the output (surface punctuation tokens) is colored in {\color{undr}green}. All arc probabilities are suppressed for readability. $\fstis$ is the start state, $\fstfs$ is the final state, $\epsilon$ denotes the empty string, and $\eoi$ denotes a special end-of-input token. The four rewriting rules at the bottom of the figure are illustrated as different paths in the PFST.}
\end{figure}

Recall from \cref{sec:nc,fig:slide} that our noisy channel is supposed to slide a 2-token window over the string of punctuation tokens, stochastically editing them as it goes.  

In our PFST implementation, each arc has the form $\circled{a}\tra{b:c}\circled{d}$, which transitions from state $a$ to state $d$ while reading an underlying punctuation token $b$ and generating a surface punctuation token $c$.  Here the state label $a$ represents the first token in the current sliding window, and the underlying token $b$ provides the second token in that window.  All surface tokens preceding $a$ have already been output by the PFST.  $a$ has not yet been output by the PFST, because it will not necessarily be part of the surface string---it might still be deleted or transposed.

Choosing to traverse this arc corresponds to choosing a particular edit to the current window contents $ab$.  After this edit, the new state $d$ will reflect the first token in the \emph{new} position of the sliding window.\footnote{Specifically, the new window contents will be $de$, where $e$ is the underlying token $e$ that follows $b$.  That token $e$ will be read by the \emph{next} arc---the arc taken from the new state $d$.}

Recall from \cref{sec:nc} that there are 4 possible edits to $ab$.  These correspond to different choices of $c$ and $d$ in $\circled{a}\tra{b:c}\circled{d}$:
\begin{itemize}
\item To allow $ab \mapsto ab$ (no change), we include an arc with $c=a$ and $d=b$.  This outputs the delayed token $a$, and then slides the window rightward so that $b$ is now the first token.
\item To allow $ab \mapsto b$ (left absorption), we include an arc with $c=\epsilon$ and $d=b$.  This is identical to the previous case, except that it chooses to skip outputting $a$, so $a$ has been deleted.
\item To allow $ab \mapsto a$ (right absorption), we include an arc with $c=\epsilon$ and $d=a$.  This is identical to the previous case, except that it is now $b$ that it skips outputting.  The first token in the sliding window therefore remains $a$.
\item To allow $ab \mapsto ba$ (transposition), we include an arc with $c=b$ and $d=a$.  This is identical to the previous case, except that it outputs $b$ \emph{before} the delayed token $a$.  We still have not output $a$, so 
  the first token in the sliding window remains $a$.
\end{itemize}
The probabilities of these 4 arcs are specified by the noisy channel parameters $\ncp$.  They must sum to 1 because our noisy channel model will choose exactly one of the 4 edits for the current sliding window $ab$.  This fact helps to ensure that our automaton is indeed a PFST, whose definition requires that the possible transitions from a given state $a$ on a given input token $b$ must have total probability of 1 \cite{cotterell14}.

We must also deal with boundary conditions, using boundary tokens $\fstisym$ and $\fstfsym$ at the start and end (respectively) of the underlying string.  
\begin{itemize}
\item The PFST starts in the special state $\fstis$, meaning that the sliding window is {\em before} the left edge of the string.  The arcs from $\fstis$ have the form $\fstis\tra{a:\epsilon}\circled{a}$ (with probability 1), which effectively edits the boundary window $\fstisym a$ by left absorption of the $\fstisym$.  
  In effect, taking the arc simply slides the window rightward to the first ``real'' position of the sliding window, discovering that its first character will be the first underlying token $a$.
\item We append the terminal token $\fstfsym$ to the underlying string.\footnote{In contrast, we did not prepend the initial token $\fstisym$ to the underlying string, but rather initialized in a state $\fstis$ that pretended that $\fstisym$ had previously been read.}
  Thus, the sliding window's final position has the form $a\fstfsym$.  The arcs that consume this token have the form $\circled{a}\tra{\fstfsym:a}\fstfs$ (with probability 1), which effectively edits the boundary window $a \fstfsym$ by right absorption of the $\fstfsym$, but with the modification that it actually emits the delayed character $a$ (which cannot undergo any further changes) and halts.
\end{itemize}

Let $\pv$ be the vocabulary of punctuation types; our PFST $\pfst$ has $|\pv|+2$ states. There is a start state  $\fstis$, a final state $\fstfs$, and the remaining $|\pv|$ states each represents a punctuation type. 
An edge $s\tra{a:b}t$ denotes a transition from state $s$ to $t$ upon reading an underlying punctuation token $a$ and generating a surface punctuation token $b$. The weight of this edge is the probability of such a transtion, which is 

The set of edges in our PFST could be enumerated as follows:
\begin{itemize}
\item $\circled{a}\ra{b:b}\circled{a},\circled{a}\ra{b:\epsilon}\circled{b},\circled{a}\ra{b:\epsilon}\circled{a},\circled{a}\ra{b:a}\circled{b}$ \hfill , for all distinct $a, b\in \pv$ 
\item $\fstis\ra{a:\epsilon}\circled{a},\circled{a}\ra{\eoi:a}\fstfs,\circled{a}\ra{a:\epsilon}\circled{a},\circled{a}\ra{a:a}\circled{a}$ \hfill , for all $a\in \pv$
\item $\fstis\ra{\eoi:\epsilon}\fstfs$ (same as the first case above but where $a=\fstfsym$ instead of $a \in \Sigma$)
\end{itemize}
\Cref{fig:pfst} illustrates the topology of our PFST with a toy vocabulary $\pv = \{\litf{a}, \litf{b}\}$. The PFST is locally normalized, because the weights of edges from a given state on the same input sum up to 1.  (See \citet{cotterell14} for a full discussion of locally normalized PFSTs.)

\paragraph{From PFST to WFSA}
In \cref{sec:algo}, we construct a weighted finite-state acceptor (WFSA) for each slot, which describes all possible underlying strings $u_i$ that can be rewritten as the surface string $x_i$ that was observed in that slot. We will explain how to obtain this WFSA.  The method is a detailed explanation of \cref{alg:line:fst} in \cref{alg:inside}, already sketched in \cref{fn:wfsa}.

First, we construct the composition $\wfst$, where $\pfst$ is the PFST as shown in yellow in \cref{fig:pfst}.  This composition extracts just the paths of $\pfst$ that would output the given surface string $\sslot_i$.  To perform this composition, we must represent the string $x_i$ as an unweighted straight-line FSA with one arc per token of $\sslot_i$.  We show this FSA in green: $\detis\ttra{\sslot_i[1]}\detcircled{1}\ttra{\sslot_i[2]}\detcircled{2}\cdots\ttra{\sslot_i[|\sslot|]}\tddc{\scriptsize|\sslot|}$.  

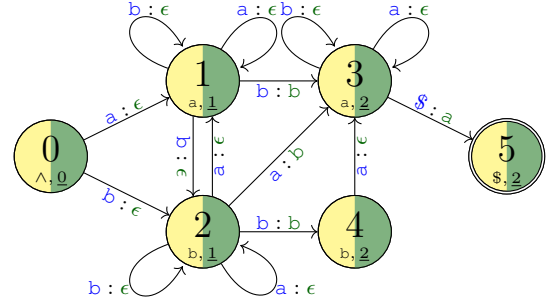
\begin{figure}[t]
\centering
\begin{tikzpicture}
\newcommand{\mynode}[5]{
    \draw node[fill=\fstclr] [#2] (#1) at #3 {};
    \fill [\detclr] (#1.north) -- (#1.south) arc[start angle=-90,delta angle=180,radius=1.25em] -- cycle;
    \node[draw=none,fill=none] at ($#3+(0, -.3)$) {\tiny #4};
    \node[draw=none,fill=none] at ($#3+(0, .1)$) {\Large #5};
    \draw node[fill=none] [#2] at #3 {};
}
\tikzstyle{every node}=[font=\scriptsize]
    [
      initial/.style={line width=1pt},
      accepting by double/.append style={line width=1pt},
      semithick,
    ]
    \mynode{i}{state}{(0,0)}{$\fstisym,\detisym$}{$0$};
    \mynode{f}{state,accepting}{(6,0)}{$\fstfsym,\tightudl{2}$}{$5$};
    \mynode{a1}{state}{(2,1)}{$\litf{a},\tightudl{1}$}{$1$};
    \path[->] (i) edge node[above=-3pt, sloped] {${\color{undb}\litf{a}}:{\color{undr}\epsilon}$} (a1);
    \path[->] (a1) edge [in=120,out=160,loop] node[above=-3pt,pos=0.7] {${\color{undb}\litf{b}}:{\color{undr}\epsilon}$} (a1);
    \path[->] (a1) edge [in=20,out=60,loop]   node[above=-3pt,pos=0.3] {${\color{undb}\litf{a}}:{\color{undr}\epsilon}$} (a1);
    \mynode{a2}{state}{(4,1)}{$\litf{a},\tightudl{2}$}{$3$};
    \path[->] (a1) edge node[below=-3pt, sloped] {${\color{undb}\litf{b}}:{\color{undr}\litf{b}}$} (a2);
    \path[->] (a2) edge [in=120,out=160,loop] node[above=-3pt,pos=0.7] {${\color{undb}\litf{b}}:{\color{undr}\epsilon}$} (a2);
    \path[->] (a2) edge [in=20,out=60,loop]   node[above=-3pt,pos=0.3] {${\color{undb}\litf{a}}:{\color{undr}\epsilon}$} (a2);
    \path[->] (a2) edge node[above=-3pt, sloped] {${\color{undb}\eoi}:{\color{undr}\litf{a}}$} (f);
    \mynode{b1}{state}{(2,-1)}{$\litf{b},\tightudl{1}$}{$2$};
    \path[->] (i) edge node[below=-3pt, sloped] {${\color{undb}\litf{b}}:{\color{undr}\epsilon}$} (b1);
    \path[->] (b1) edge [out=300,in=340,loop] node[right=-3pt] {${\color{undb}\litf{a}}:{\color{undr}\epsilon}$} (b1);
    \path[->] (b1) edge [out=200,in=240,loop] node[left=-3pt,pos=.465] {${\color{undb}\litf{b}}:{\color{undr}\epsilon}$} (b1);
    \path[->] (a1.255) edge node[below=-3pt,sloped] {${\color{undb}\litf{b}}:{\color{undr}\epsilon}$} (b1.105);
    \path[->] (b1.75) edge node[below=-3pt,sloped] {${\color{undb}\litf{a}}:{\color{undr}\epsilon}$} (a1.-75);
    \path[->] (b1) edge node[below=-3pt, sloped] {${\color{undb}\litf{a}}:{\color{undr}\litf{b}}$} (a2);
    \mynode{b2}{state}{(4,-1)}{$\litf{b},\tightudl{2}$}{$4$};
    \path[->] (b1) edge node[above=-3pt] {${\color{undb}\litf{b}}:{\color{undr}\litf{b}}$} (b2);
    \path[->] (b2) edge node[below=-3pt,sloped] {${\color{undb}\litf{a}}:{\color{undr}\epsilon}$} (a2);
\end{tikzpicture}
\caption{\label{fig:post_compose} The WFST obtained by composing the yellow PFST $\pfst$ in \cref{fig:pfst} with the green straight-line FSA $\detis\ttra{\litf{b}}\detcircled{1}\ttra{\litf{a}}\detdcircled{2}$  that accepts $\sslot_i = \litf{ba}$.  The states are indexed from $0$ (the initial state) to $5$ (the final state). The bottom of each state shows the identifiers of the yellow and green states that it combines.  Each arc is copied, along with its labels and weight, from a corresponding arc in \cref{fig:pfst}.  Only states that are accessible from the initial state are shown; arc weights are suppressed for readability.
}
\end{figure}

The composition $\wfst$ is illustrated in \cref{fig:post_compose}. Each state in the composition has the form $\tc{y,\hspace{-2pt}z}$, where $y$ is some yellow state identifier in $\pfst$ and $z$ is some green state identifier in the straight-line FSA for $\sslot_i$.   Thus, we depict it in \cref{fig:post_compose} as a yellow/green state.
In other words, the state space of $\wfst$ consists of the Cartesian product of the PFST states and the straight-line FSA states.
The edge $\tc{y,\hspace{-2pt}z}\ttra{s}\ttc{\hspace{-2.5pt}y'\hspace{-4pt},\hspace{-2pt}z'}$ exists if and only if  $\tc{y}\ttra{s:t}\tc{y'}$ exists in $\pfst$ and $\tc{z}\ttra{t}\tc{z'}$ exists in $\sslot$, with the edge weight inherited from the former.
Note that the result of composition is a WFST rather than a PFST, since the arc weights are no longer guaranteed to be locally normalized.

Finally, to obtain the desired WFSA that describes the possible underlying strings $\uslot_i$ that could have yielded $\sslot_i$, we project the WFST onto its domain (input).  This is a simple matter of dropping the output (which follows the colon) from each arc in the WFST of \cref{fig:post_compose}.  The weights are retained.

\newcommand{\pra}[1]{\hspace{-3pt}\mathrel{\raisebox{-3pt}{$\stackrel{#1}{\rightsquigarrow}$}}\hspace{-3pt}}

\end{document}